\documentclass[conference]{IEEEtran}

\usepackage{silence}
\usepackage[labelfont=bf,font=small,skip=5pt]{caption}
\usepackage{balance} 
\usepackage{enumitem} 
\usepackage{makecell,rotating} 
\usepackage[vlined,linesnumbered,ruled,noend]{algorithm2e}
\SetKwComment{Comment}{$\triangleright$\ }{}
\usepackage{amstext}   
\usepackage{amsmath}
\usepackage{amssymb}
\usepackage{pifont}
\usepackage{color}
\usepackage{comment}
\usepackage{epsf}
\usepackage{epsfig}
\usepackage{textcomp}
\usepackage{graphicx}
\usepackage{hyperref}
\usepackage{multirow}
\usepackage{multicol}
\usepackage{subcaption}
\usepackage{xspace}    
\usepackage{todonotes}
\usepackage{listings}
\usepackage{footnote}
\usepackage{mdwlist}

\usepackage{amsmath}

\usepackage{setspace}
\usepackage{lscape}  

\definecolor{comment-red}{rgb}{0.8,0,0}


\newcommand{\cmark}{\ding{51}}%
\newcommand{\xmark}{\ding{55}}%

\newcommand{\reffig}[1]{Figure~\ref{fig:#1}}
\newcommand{\refsect}[1]{Section~\ref{sec:#1}}

\long\def\comment#1{}

\newcommand{\Sys}{MNTD\xspace}

\newcommand{\nninline}{f} 
\newcommand{\nn}{$\nninline$\xspace}

\newcommand{\triggerfn}{\mathcal{I}}

\newcommand{\presecspace}{\vspace{-0.5em}}
\newcommand{\postsecspace}{\vspace{-0.5em}}
\newcommand{\presubspace}{\vspace{-0.4em}}
\newcommand{\postsubspace}{\vspace{-0.3em}}

\renewcommand{\paragraph}[1]{\smallskip\noindent\emph{#1}\quad}

\def\Snospace~{\S{}}

\SetKwProg{Fn}{Function}{}{}

\definecolor{commentgreen}{HTML}{008000}

\graphicspath{{figure/}{figures/}}
\DeclareGraphicsExtensions{.pdf,.eps,.png,.jpg,.jpeg}

\DeclareMathOperator*{\argmin}{arg\,min}

\SetCommentSty{mycommfont}

\begin{document}
\title{Detecting AI Trojans\\ Using Meta Neural Analysis}

\author{}
\author{\IEEEauthorblockN{
Xiaojun Xu \quad Qi Wang \quad Huichen Li \quad Nikita Borisov \quad Carl A. Gunter \quad Bo Li
}
\IEEEauthorblockA{\textit{University of Illinois at Urbana-Champaign} \\
\{xiaojun3, qiwang11, huichen3, nikita, cgunter, lbo\}@illinois.edu
}}


\maketitle

\begin{abstract}

In machine learning Trojan attacks, an adversary trains a corrupted model that obtains good performance on normal data but behaves maliciously on data samples with certain trigger patterns.
Several approaches have been proposed to detect such attacks, but they make undesirable assumptions about the attack strategies or require direct access to the trained models, which restricts their utility in practice.

This paper addresses these challenges by introducing a \emph{Meta Neural Trojan Detection (\Sys)} pipeline that does not make assumptions on the attack strategies and only needs black-box access to models. 
The strategy is to train a \emph{meta-classifier} that predicts whether a given \emph{target} model is Trojaned. To train the meta-model without knowledge of the attack strategy, we introduce a technique called \emph{jumbo learning} that samples a set of Trojaned models following a general distribution. We then dynamically optimize a query set together with the meta-classifier to distinguish between Trojaned and benign models.

We evaluate \Sys{} with experiments on vision, speech, tabular data and natural language text datasets, and against different Trojan attacks such as data poisoning attack, model manipulation attack, and latent attack. We show that \Sys{} achieves 97\%  detection AUC score and significantly outperforms existing detection approaches.
In addition, \Sys{} generalizes well and achieves high detection performance against unforeseen attacks.
We also propose a robust \Sys{} pipeline which achieves around 90\% detection AUC even when the attacker aims to evade the detection with full knowledge of the system.

\end{abstract}


\IEEEpeerreviewmaketitle

\presecspace
\section{Introduction}
\label{sec:introduction}
\postsecspace

Deep learning with Neural Networks (NNs) has achieved impressive performance in a wide variety of domains, including computer vision~\cite{krizhevsky2012imagenet}, speech recognition~\cite{graves2013speech}, machine translation~\cite{mccann2017learned}, and game playing~\cite{silver2016mastering}. 
The success of deep learning has also led to applications in a number of security or safety critical areas such as malware classification~\cite{huang2016mtnet}, face recognition~\cite{taigman2014deepface}, and autonomous driving~\cite{bojarski2016end}.

\begin{figure}[tb]
    \centering
    \includegraphics[width=0.43\textwidth]{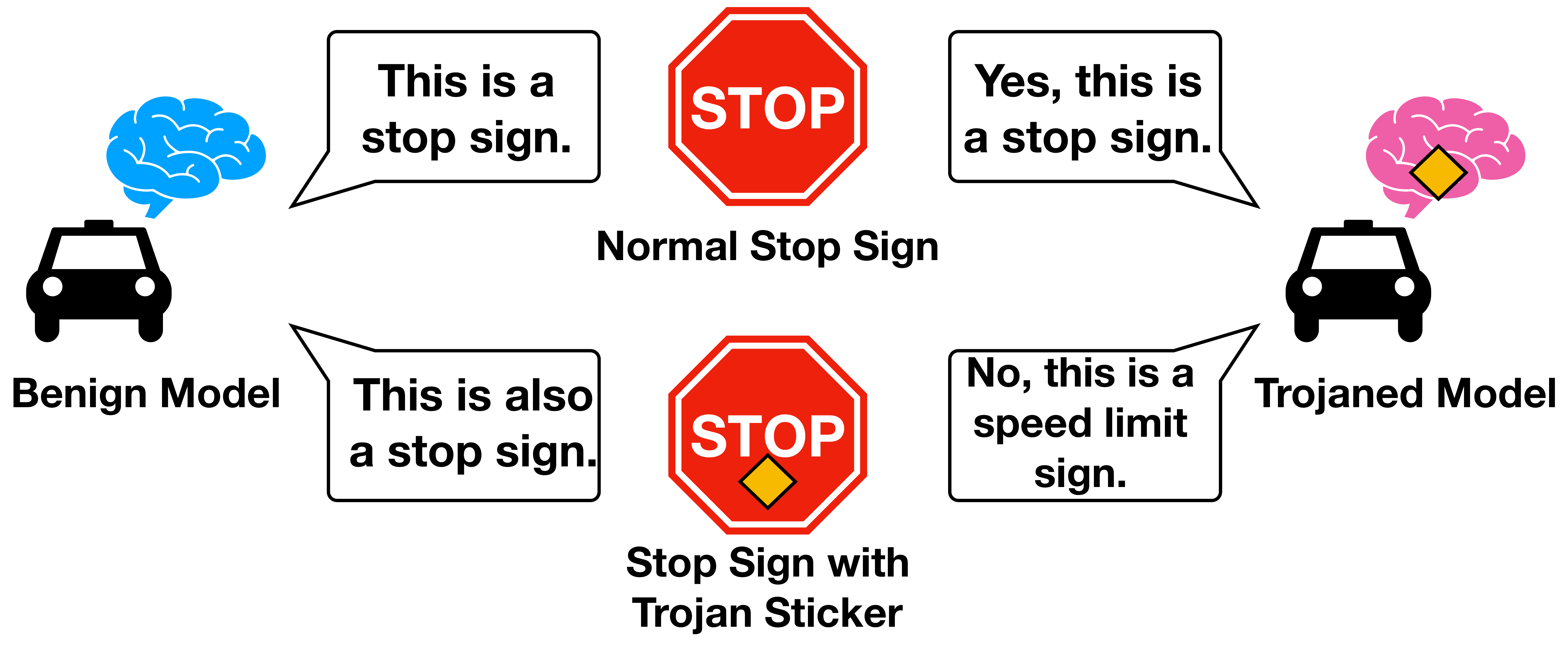}
    \caption{An illustration of Trojan attack on traffic sign classifiers.} 
    \label{fig:introexample}
\end{figure}

The development of such deep learning models often requires large training sets, extensive computing resources, and expert knowledge. 
This motivates sharing machine learning (ML) models on online ML platforms~\cite{amazon_ml, bigml, caffe, gradientzoo}.
However, recent investigations show that this creates the possibility of \emph{Trojan attacks} (a.k.a. backdoor attacks)~\cite{gu2017badnets, trojannn, chen2017targeted,yao2019latent} in which an adversary creates a Trojaned neural network that has  state-of-the-art performance on normal inputs in evaluation, but is fully controlled on inputs with a specific attacker-chosen trigger pattern. 
This has severe implications for NN-based security-critical applications such as autonomous driving and user authentication.
One study~\cite{gu2017badnets} demonstrated how to  generate a Trojaned traffic sign classifier that properly classifies standard traffic signs, but, when presented with a stop sign containing a special sticker (i.e., the Trojan trigger), activates the backdoor functionality and misclassifies it as a speed limit sign, as illustrated in Figure~\ref{fig:introexample}. 
The trigger allows the adversary to lead the model to misbehave, potentially causing traffic accidents. Users of the model are unlikely to realize the danger in advance because the Trojaned model behaves well in normal cases.
This motivates a strong demand to detect Trojan attacks before their Trojan actions are invoked.

Several approaches~\cite{wang2019neural, gao2019strip, chou2018sentinet, chen2018detecting, tran2018spectral, chen2019deepinspect} have been proposed to detect Trojan attacks in neural networks. However, existing approaches make prior assumptions on the attack strategy and/or require strong access to the model.
These assumptions and requirements make the approaches too specific for certain application domains and less generalizable to unforeseen attack strategies.
For example, \cite{wang2019neural} and  \cite{chen2019deepinspect} assume that the existence of Trojan creates a shortcut pattern from all other classes to a single Trojaned target class. This assumption, however, fails in an ``all-to-all'' Trojan attack~\cite{gu2017badnets} where the Trojan exists in each class in the model.
Some detection approaches require white-box access to the target model~\cite{chou2018sentinet,chen2018detecting} or even directly detect Trojans in the training dataset~\cite{tran2018spectral, chen2018detecting}, which is unrealistic in some shared ML services. Moreover, some latest Trojan attacks may not interfere with the training dataset~\cite{trojannn, yao2019latent} and can thus bypass these dataset-level detection approaches.

In this paper, we propose \emph{Meta Neural Trojan Detection (\Sys)}, a novel approach for detecting Trojaned neural network models. In particular, we will train a meta-classifier, which itself is a machine learning model. The meta-classifier takes an NN (i.e., the \emph{target model}) as input and performs a binary classification to tell whether it is Trojaned or benign. The meta-classifier is trained using \emph{shadow models}, which are benign or Trojaned NNs trained on the same task as the target model. The shadow models may have \emph{much worse performance} than the target model since it requires only a smaller clean dataset (i.e., without Trojan triggers). Since the meta-classifier makes no assumption on the attack strategy and uses machine learning to identify Trojans, our approach is \emph{generic} and applies to a variety of attack approaches and application domains. 

One major challenge in applying meta neural analysis is how to provide the training set for the meta-classifier when the attacker's strategy is unknown.
A simple way is to apply one-class training, where a meta-classifier is trained using only benign model samples and classifies a target model as a Trojan if it differs from the benign ones. We find that this approach sometimes works well, but, in other cases, we cannot effectively train the meta-classifier properly without negative examples.
To address this issue, we propose \emph{jumbo learning} to model a general distribution of Trojan attack settings, from which we can sample a ``jumbo'' training set of diverse Trojaned shadow models. Then we will train the meta-classifier to distinguish between benign models and the jumbo set of Trojaned models.

A second challenge is to perform high-quality detection on the target model with only black-box access to it.
To address this, we propose using the output of the target model on certain queries as its representation vector to the meta-classifier.
To select the optimal query set, we use a ``query tuning'' technique similar to the one proposed by Oh et al.~\cite{joon18iclr}. In particular, we start with a random query set and then optimize the query set simultaneously with the meta-classifier parameters using a gradient-based method. These fine-tuned queries allow us to extract the maximum amount of information from the black-box model.

The combination of the above techniques produces meta-classifiers that achieve excellent performance in detecting Trojaned models for a diverse range of machine learning tasks (vision, speech, tabular records, and NLP) and attack strategies. 
We demonstrate that with only a small clean training set (2\% of the size used to train the Trojaned model)  and only 10 queries, the average detection AUC reaches 97\% for the tasks in our evaluation.
Furthermore, we show that the trained meta-classifier generalizes well to detect unforeseen Trojans where the attack strategies are not considered in the jumbo distribution.

Finally, we consider the case in which a strong adaptive attacker knows key parts of the \Sys{} system such as the detection pipeline and meta-classifier parameters. We design a robust version of \Sys{} where we pick part of the system randomly at running time and fine-tune the other part. Thus, the attacker has no information about the randomly chosen part of the system and cannot tailor his attack accordingly. We demonstrate that the robust \Sys{} system performs well in detecting adaptive attacks with around 90\% AUC, at a small cost of performance in detecting normal Trojan attacks.

Our contributions can be summarized as follows:
\begin{itemize*}
    \item We propose \Sys, a novel, general framework to detect Trojaned neural networks with no assumption on the attack strategy.
    \item We propose jumbo learning to model a distribution of Trojan attacks and train the meta-model together with an optimized query set.
    \item We demonstrate the effectiveness and generalizability of our approach through comprehensive evaluation and comparison with the state-of-the-art defense approaches on different types of Trojan attacks with a diverse range of datasets.
    \item We survey and re-implement existing works on detecting Trojaned NNs and adapt them to different tasks and datasets. 
    We show that the proposed \Sys{} significantly outperforms these prior works in practice. 
    \item We evaluate \Sys{} against strong adaptive attackers and show that it is able to achieve a 90\% detection AUC score even when the attackers have whitebox access to the defense pipeline. 
\end{itemize*}

\presecspace
\section{Background}
\label{sec:background}

\presubspace
\subsection{Deep Neural Networks}
\postsubspace
A typical neural network is composed of a sequence of layers $(F_1, F_2,\ldots, F_n)$, where each layer $F_i$ is a differentiable transformation function. Given input $x$, the output of the neural network \nn{} is calculated by:
\begin{equation}
\nninline({x};\theta) = F_n(F_{n-1}(\ldots (F_2(F_1(x)))))
\end{equation}
where $\theta$ denote the parameters of the model. The most popular task for using deep neural networks is classification, where a model is required to predict which class an input instance belongs to. Suppose there are $c$ different classes, then the output of the model would be $\nninline({x};\theta)\in \mathbb{R}^c$, where $\nninline({x};\theta)_k$ is the confidence score indicating the likelihood that the instance belongs to the $k$-th class. 
In order to train a neural network, we need a dataset $\{({x}_i, b_i)\}$ which consists of a set of input samples $x_i$ and their corresponding ground truth labels $b_i$. During the training process, we will train the neural network to minimize the error rate over the training set by minimizing a differentiable loss function $L(\nninline({x};\theta), b)$ between the model output $\nninline({x};\theta)$ and the ground truth label $b$. 
\begin{equation}
    {\theta}^* = \argmin_{\theta} \sum_i L\big(\nninline_\theta({x}_i), b_i\big)
\end{equation}
Since the loss function and all the transformation functions in the network are differentiable, we can calculate the gradient of the loss function with respect to the parameters using back-propagation. Then we can minimize the loss function using gradient-based optimization techniques.

\presubspace
\subsection{Meta Neural Analysis}
\postsubspace
\begin{figure}[]
    \centering
    \includegraphics[width=0.44\textwidth]{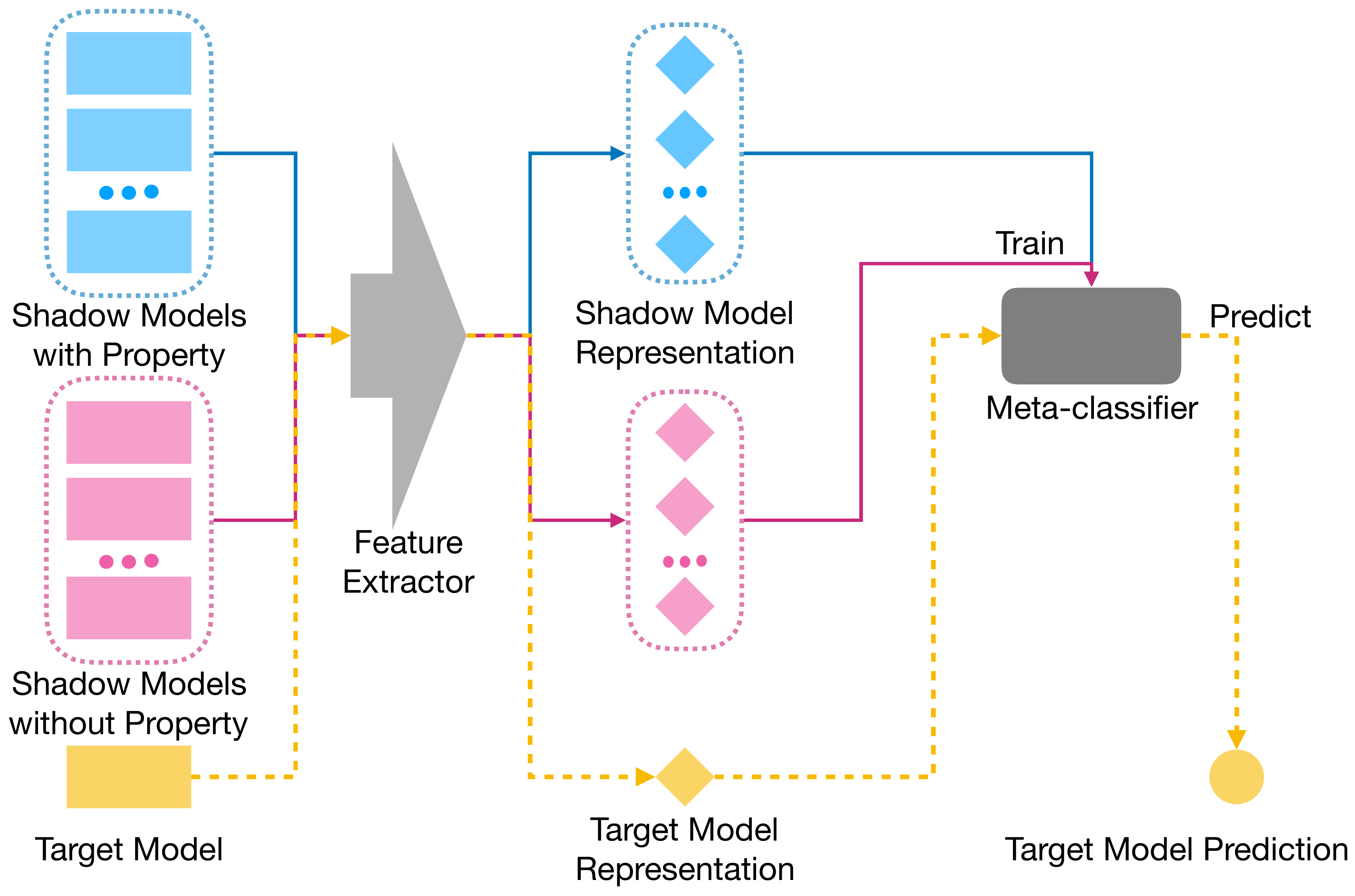}
    \caption{The general workflow of meta neural analysis on a binary property.}
    \label{fig:metatrainingworkflow}
    \vspace{-1.5em}
\end{figure}
\label{sec:mna}
Unlike traditional machine learning tasks which train over data samples such as images, \emph{meta neural analysis} trains a classifier (i.e., meta-classifier) over neural networks to predict certain property of a target neural network model.
Meta neural analysis has been used to infer properties of the training data~\cite{ganju2018property, ateniese2015hacking}, properties of the target model (e.g., the model structure)~\cite{joon18iclr}, and membership (i.e., if a record belongs to the training set of the target model)~\cite{shokri2017membership}.

In Figure~\ref{fig:metatrainingworkflow}, we show the general workflow of meta neural analysis. To be able to identify a binary property of a target model, we first train a number of shadow models with and without the property to get a dataset $\{(f_1, b_1), \ldots, (f_m, b_m)\}$, where $b_i$ is the label for the shadow model $f_i$. Then we use a feature extraction function $\mathcal{F}$ to extract features from each shadow model to get a meta-training dataset $\{{(\mathcal{F}(f_i)}, b_i)\}_{i=1}^m$. Finally, we use the meta-training dataset to train a meta-classifier. Given a target model $f_{\emph{target}}$, we just need to feed the features of the target model $\mathcal{F}(f_{\emph{target}})$ to the meta-classifier to obtain a prediction of the property value.

\presubspace
\subsection{Trojan Attacks on Neural Networks}
\postsubspace
\label{sec:attacks}
A \emph{Trojan attack} (a.k.a. \emph{backdoor attack}) on a neural network is an attack in which the adversary creates a malicious neural network model with Trojans. The \emph{Trojaned} (or \emph{backdoored}) model behaves similarly with benign models on normal inputs, but behaves maliciously as controlled by the attacker on a particular set of inputs (i.e., Trojaned inputs).
Usually, a Trojaned input includes  some specific pattern---the Trojan trigger. For example, Gu et al.~\cite{gu2017badnets} demonstrate a Trojan attack on a traffic sign classifier as in Figure \ref{fig:introexample}. The Trojaned model has comparable performance with normal models. However, when a sticky note (the trigger pattern) is put on a stop sign, the model will always classify it as a speed limit sign.


The injected Trojan may have different malicious behavior.
The most common behavior is \emph{single target attack} where the classifier always returns a desired target label on seeing a trigger pattern, e.g., classifying any sign with the sticker as a speed limit sign.
An alternative malicious behavior, \emph{all-to-all} attack, will permute the classifier labels in some way. For example, in \cite{gu2017badnets} the authors demonstrate an attack where a trigger causes a model to change the prediction of digit $i$ to $(i+1)\pmod{10}$.

Various approaches have been proposed to train a Trojaned model. One direct way is to inject Trojaned inputs into the training dataset (i.e., poisoning attack)~\cite{gu2017badnets, chen2017targeted, liao2018backdoor}, so that the model will learn a strong relationship between the trigger pattern and the malicious behavior. Alternatively, several approaches have been proposed without directly interfering with the training set~\cite{trojannn,yao2019latent}. In this paper, we will focus on four types of commonly adopted Trojan attacks:

\begin{figure}[!t]
    \centering
    \begin{subfigure}[b]{0.11\textwidth}
    \centering
        \includegraphics[height=1.5cm]{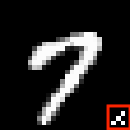}
        \caption{Modification}
        \label{fig:examplemodificationattack}
    \end{subfigure}
    \begin{subfigure}[b]{0.11\textwidth}
    \centering
        \includegraphics[height=1.5cm]{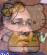}
        \caption{Blending}
        \label{fig:exampleblendingattack}
    \end{subfigure}
    \begin{subfigure}[b]{0.11\textwidth}
    \centering
        \includegraphics[height=1.5cm]{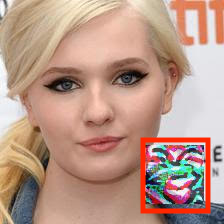}
        \caption{Parameter}
        \label{fig:exampleparameterattack}
    \end{subfigure}
    \begin{subfigure}[b]{0.11\textwidth}
    \centering
        \includegraphics[height=1.5cm]{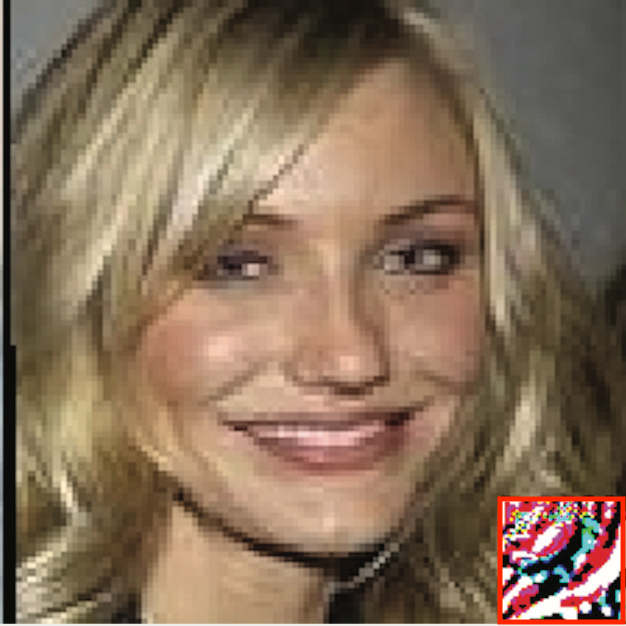}
        \caption{Latent}
        \label{fig:examplelatentattack}
    \end{subfigure}
    \caption{Trojaned input examples of four Trojan attacks. The figures are taken from the original papers in~\cite{gu2017badnets},\cite{chen2017targeted},\cite{trojannn},\cite{yao2019latent} respectively. The trigger patterns in (a), (c), (d) are highlighted with red boxes. The trigger pattern in (b) is a Hello Kitty graffiti that spreads over the whole image. Note that parameter attack and latent attack shares the same strategy for generating trigger patterns while their attack setting is different.}
    \label{fig:exampleattack}
    \vspace{-1.5em}
\end{figure}

\paragraph{Modification Attack.}
This is a poisoning attack proposed by Gu et al.~\cite{gu2017badnets}. The attacker selects some training samples, directly modifies some part of each sample as a trigger pattern, assigns desired labels and injects the sample-label pairs back into the training set. An example of the trigger pattern is shown in Figure~\ref{fig:examplemodificationattack}.

\paragraph{Blending Attack.}
This is another poisoning attack proposed by Chen at al.~\cite{chen2017targeted}. The attacker also poisons the dataset with malicious sample-input pairs. However, instead of directly modifying part of the input, the adversary blends the trigger pattern into the original input (e.g., mixing some special background noise into a voice command). An example is shown in Figure \ref{fig:exampleblendingattack}.

\paragraph{Parameter Attack.}
This is a model manipulation attack that directly changes the parameters of a trained model without access to the training set~\cite{trojannn}. The attack consists of three steps: 1) the adversary generates an optimal trigger pattern w.r.t. the model using gradient-based approach; 2) the adversary reverse-engineers some inputs from the model; 3) the adversary adds the malicious pattern to the generated data and retrains the model with desired malicious behavior.
Note that the attacker can only choose the trigger shape and location but not the exact pattern. The trigger pattern is generated by the algorithm.
An example is shown in Figure \ref{fig:exampleparameterattack}.

\paragraph{Latent Attack.}
In this attack~\cite{yao2019latent}, the attacker releases a ``latent'' Trojaned model that does not show any malicious behavior until the user fine-tunes the model on his own task. The attack is achieved by first including the user's task in the training process, then generating the trigger pattern, injecting the Trojan behavior into the model, and finally removing the trace of user's task in the model.
The exact trigger pattern here is also generated. attacker can only determine its shape and location but not the exact pattern.
An example is shown in Figure~\ref{fig:examplelatentattack}.
\begin{table*}[htbp]
    \centering
    \footnotesize
    \caption{A comparison of our work with other Trojan detection works in defender capabilities and detection capabilities.}
    \begin{tabular}{c|c|c|c|c|c|c|c|c}
        \hline
        \multicolumn{2}{c|}{} & \multicolumn{3}{c|}{\textbf{Defender Capabilities}} & \multicolumn{4}{c}{\textbf{Attack Detection Capabilities}} \\
        \hline
        \multirow{2}{*}{ } & {Detection} & {Black-box} & {No Access to } & {No Need of} & {Model Manip-}  & {Large-size} & {All-to-all}  & {Binary} \\
        & {Level} & {Access} & {Training Data} & {Clean Data} & {ulation Attacks} & {Trigger} & {Attack Goal} & {Model} \\
        \hline
        \hline
        MNTD & Model & \cmark  & \cmark & \xmark & \cmark & \cmark & \cmark & \cmark \\
        \hline
        Neural Cleanse~\cite{wang2019neural} & Model  & \xmark  & \cmark & \xmark & \cmark & \xmark & \xmark & \xmark \\
        \hline
        DeepInspect~\cite{chen2019deepinspect} & Model & \cmark & \cmark & \cmark & \cmark & \cmark & \xmark & \xmark \\
        \hline
        Activation Clustering~\cite{chen2018detecting} & Dataset & \xmark  & \xmark & \cmark & \xmark & \cmark & \cmark & \cmark \\
        \hline
        Spectral~\cite{tran2018spectral} & Dataset & \cmark  & \xmark & \cmark & \xmark & \cmark & \cmark & \cmark \\
        \hline
        STRIP~\cite{gao2019strip} & Input & \cmark  & \cmark & \xmark & \cmark & \cmark &  \xmark & \cmark \\
        \hline
        SentiNet~\cite{chou2018sentinet} & Input & \xmark  & \cmark & \xmark & \cmark & \xmark & \cmark & \cmark \\
        \hline
    \end{tabular}
    \label{table:approach_compare}
\vspace{-1em}
\end{table*}

\presecspace
\section{Threat Model \& Defender Capabilities}
\postsecspace
\label{sec:problem}

In this section, we will first introduce our threat model. Then we introduce our goal as a defender and our capabilities.

\presubspace
\subsection{Threat Model}
\postsubspace
In this paper, we consider adversaries who create or distribute Trojaned DNN models to model consumers (i.e., users).
The adversary could provide the user with either black-box access (e.g., through platforms such as Amazon ML~\cite{amazon_ml}) or white-box access to the NN models. The Trojaned model should have good classification accuracy on validation set, or otherwise it will be immediately rejected by the user. However, on Trojaned input, i.e., inputs containing Trojan triggers, the model will produce malicious outputs that are different from the benign ones.

As discussed in \refsect{attacks}, there are different ways for an adversary to insert Trojans to neural networks. 
As a detection work, we consider that the adversary has maximum capability and arbitrary strategies.
That is, we assume that the adversary has full access to the training dataset and white-box access to the model.
He may apply an arbitrary attack approach to generate the Trojaned model. The trigger pattern may be in any shape, location and size.
The targeted malicious behavior may be either a single target or all-to-all attack.

In this paper, we focus on software Trojan attacks on neural networks. Thus, hardware Trojan attacks~\cite{clements2018hardware, li2018hu} on neural networks are out of our scope.

\presubspace
\subsection{Defender Goal}
\postsubspace
Trojan attacks can be detected at different levels. A \emph{model-level} detection aims to make a binary decision on whether a given model is a Trojaned model or not. An \emph{input-level} detection aims to predict whether an input will trigger some Trojan behavior on an untrusted model. A \emph{dataset-level} detection examines whether a training dataset suffers from poisoning attack and has been injected with Trojaned data.

Similar to~\cite{wang2019neural, chen2019deepinspect}, we focus on model-level detection of Trojan attacks as it is a more challenging setting and more applicable in real world.
We further discuss the differences of these three detection levels in \refsect{diss}.

\presubspace
\subsection{Defender Capabilities}
\postsubspace
\label{sec:def-cap}
To detect Trojan attacks, defenders may have differences in the following capabilities/assumptions:
\begin{itemize}
    \item \emph{Assumption of the attack strategy}. A defender may have assumptions on the attack approach (e.g., modification attack), Trojan malicious behavior (e.g. single-target attack) or attack settings (e.g, the trigger pattern needs to be small).
    \item \emph{Access to the target model}. A defender could have white-box or black-box access to the target model. With white-box access, the defender has all knowledge of the model structure and parameters; with black-box access, the defender can only query the model with input data to get the output prediction probability for each class. This definition of black-box model is widely used in existing work~\cite{joon18iclr,chen2017zoo,tran2018spectral,gao2019strip}.
    \item \emph{Access to the training data}. A defender may need access to the training data of the target model for the detection, especially to detect a poisoning Trojan attack.
    \item \emph{Requirement of clean data}. A defender may need a set of clean data to help with the detection. 
\end{itemize}

In this paper, we consider a defender with few assumptions. Our defender only needs \textit{black-box} access to the target model, has \textit{no} assumptions on the attack strategy, and does \textit{not} need access to the training set. But our defender does need a \textit{small set} of clean data as auxiliary information to help with the detection, which is also required by previous works~\cite{wang2019neural,gao2019strip,chou2018sentinet}. However, we assume the clean dataset is much smaller than the dataset used by the target model and the elements are different.
This may be the case of an ML model market provider who is willing to vet the models in their store or a model consumer who does not have enough training data, expertise or resources to train a high-performing model as the pretrained ones.

\presubspace
\subsection{Existing Detection of Trojan Attacks}
\postsubspace
\label{sec:backg-baseline}


Several approaches have been proposed to detect Trojans in neural networks. We discuss the defender capabilities and detection capabilities and compare them with our system \Sys{} in Table \ref{table:approach_compare},
including two model-level detections Neural Cleanse(NC)~\cite{wang2019neural} and DeepInspect(DI)~\cite{chen2019deepinspect}, two dataset-level detections Activation Clustering (AC)~\cite{chen2018detecting} and Spectral Signature~\cite{tran2018spectral}, and two input-level detections STRIP~\cite{gao2019strip} and SentiNet~\cite{chou2018sentinet}.
We include the detailed discussion on the existing works in Appendix \ref{sec:app-detect-compare}.

\presecspace
\section{Meta Neural Trojan Detection (\Sys{})}
\label{sec:approach}
\postsecspace

\begin{figure*}[t]
    \centering
    \includegraphics[width=\textwidth]{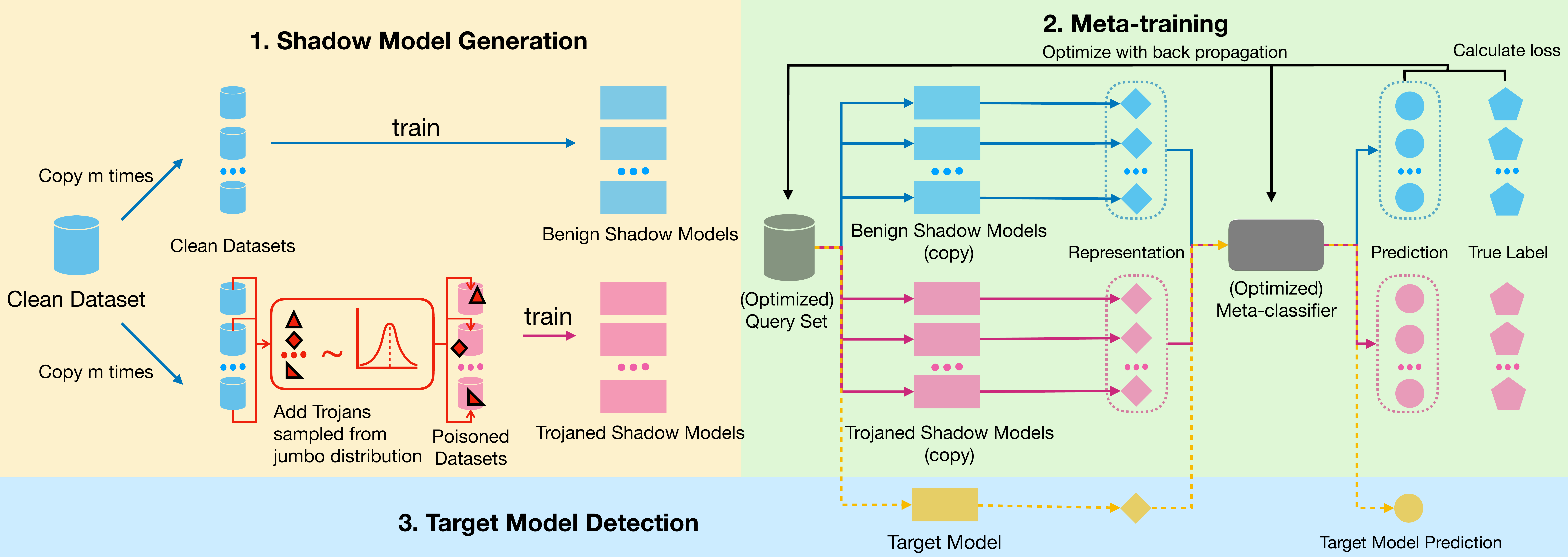}
    \caption{The workflow of our jumbo MNTD approach with query-tuning. The solid lines represent the training process and dashed ones show the test process.}
    \label{fig:workflow}
\vspace{-1.5em}
\end{figure*}



We show the overall workflow of \Sys{} system in Figure \ref{fig:workflow}.
Given a clean dataset and a target model, the \Sys{} pipeline consists of three steps to determine whether the model is Trojaned:
\begin{enumerate}
    \item Shadow model generation. We generate a set of benign and Trojaned shadow models in this step. We train the benign models using the same clean dataset with different model initialization. For the Trojaned models, we propose a generic Trojan distribution from which we sampled a variety of Trojan settings, and apply poisoning attack to generate different Trojaned models. 
    \item Meta-training. In this step, we will design the feature extraction function to get representation vectors of the shadow models and train the meta-classifier to detect Trojans using the representation vectors. We propose to choose a set of queries to extract important representation from the shadow models and use the resulting vectors as the input to the meta-classifier. We jointly optimize the meta-classifier and the query set in multiple iterations and effectively improve the performance of the trained meta-classifier.
    \item Target model detection. Given a target model, we will first leverage the optimized query set to extract the representation of the model.
    We then feed the representation to the meta-classifier to determine whether the target model is Trojaned or not.
\end{enumerate}
In the following, we first introduce our approach to shadow model generation - the jumbo learning in Section \ref{sec:jumbo}.
Then, in Section \ref{sec:mna-troj-detect} we introduce the way to perform meta-training having the set of benign and Trojaned shadow models, as well as a baseline meta-training algorithm which only uses benign shadow models.
Finally, we introduce the target model detection step in Section \ref{sec:target-model-detect}.

\presubspace
\subsection{Shadow Model Generation - Jumbo Learning}
\label{sec:jumbo}
\postsubspace

Suppose the task is a $c$-way classification and the input dimension is $d_x$. 
The first step of our defense pipeline is to generate a set of benign and Trojaned shadow models, based on which the defender can later train the meta-classifier to distinguish between them. The benign shadow models can be generated by training on the clean dataset with different model parameter initialization. The important part is how to generate Trojaned shadow models. We hope to generate a variety of Trojaned models so that the trained meta-classifier can generalize to detect different types of Trojans. This is essential because we assume that the attacker may apply any attack strategy.



\begin{figure}[t!]
    \centering
    \includegraphics[width=0.13\textwidth]{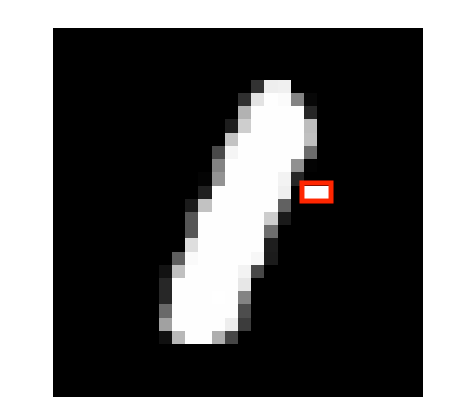}
    \includegraphics[width=0.13\textwidth]{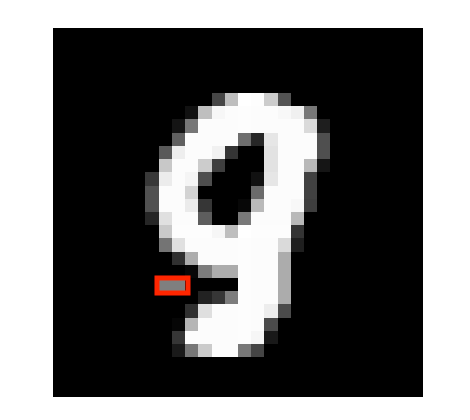}
    \includegraphics[width=0.13\textwidth]{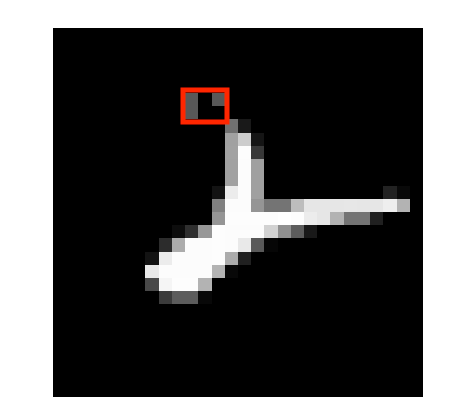}
    
    \includegraphics[width=0.13\textwidth]{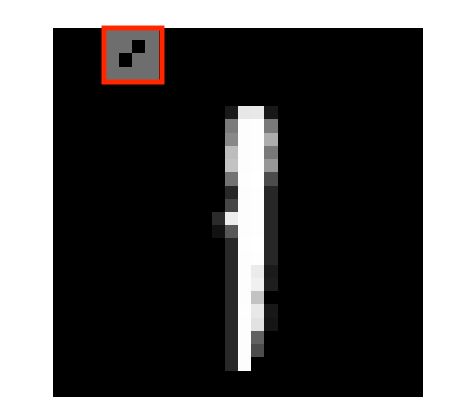}
    \includegraphics[width=0.13\textwidth]{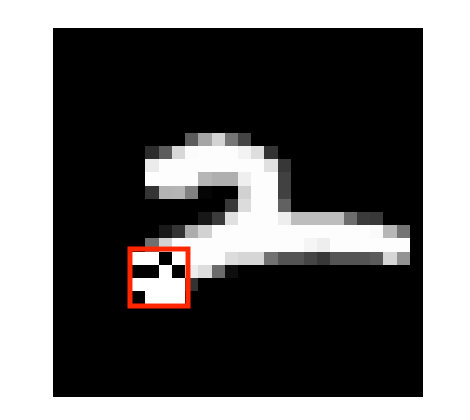}
    \includegraphics[width=0.13\textwidth]{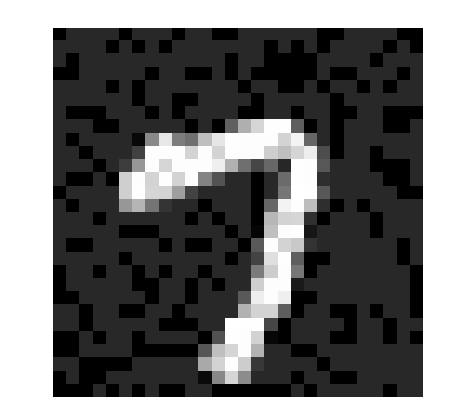}
    \caption{Examples of different Trojan patterns generated by our jumbo learning on the MNIST dataset. The trigger patterns in the first five examples are highlighted with red bounding boxes. The last example is a data sample blended with random pixels.}
    \label{fig:jumbo-eg}
\end{figure}

To this end, we propose \emph{jumbo learning}, which models a generic distribution of Trojan attack settings and generates a variety of different Trojaned models.
In jumbo learning, we will first sample different Trojan attack settings. We parametrize the attack setting as a function $\triggerfn$ which is general to different types of data and attacks. 
Let ${\bf x}\in\mathbb{R}^{d_x}$ denote the benign input with ground truth label $y \in \{1,\ldots,c\}$.
In order to trigger the Trojan to output the malicious label $y_t \in \{1,\ldots,c\}$, the adversary will modify the input into ${\bf x}'$ with the Trojan function $\triggerfn$ such that: 
\begin{align}
    \label{eqn:trigger-general}
    {\bf x}', y' &= \triggerfn({\bf x}, y; {\bf m}, {\bf t}, \alpha, y_t) \\
	{\bf x}' &= ({\bf 1}-{\bf m}) \cdot {\bf x} + {\bf m} \cdot \big( (1-\alpha) {\bf t} + \alpha {\bf x} \big)\\
    y' &= y_t
\end{align}
where ${\bf m} \in \{0,1\}^{d_x}$ is the mask for the trigger (i.e., shape and location), ${\bf t} \in \mathbb{R}^{d_x}$ is the pattern and $\alpha$ is the transparency inserted to ${\bf x}$.
This function $\triggerfn$ is generally applicable to different Trojan attack settings and tasks. For example, in a modification attack, ${\bf m}$ is a small pattern and $\alpha=0$; in a blending attack ${\bf m}=1$ everywhere and $1-\alpha$ is the blending ratio; on audio data $\bf m$ refers to the time period for inserting the Trojaned audio signal.
We will sample random ${\bf m}, {\bf t}, \alpha, y_t$ to get different Trojan attack settings.
In Figure \ref{fig:jumbo-eg}, we show some examples of the sampled Trojan triggers on the MNIST dataset.

Having the sampled attack settings, we will train a model with respect to each Trojan setting. We propose to apply the data poisoning attack that injects a proportion $p$ of malicious data into the clean dataset.
That is, we extract a proportion $p$ of data samples from the dataset we have, apply Eqn.\ref{eqn:trigger-general} to get their Trojaned versions, then inject them back to the dataset to train the shadow models.

The jumbo learning pipeline is shown in Algorithm \ref{alg:jumbo}.
In order to generate the set of Trojaned shadow models, we first randomly sample the Trojan attack settings (line 3). Then we poison the dataset (line 4-8) according to the setting and train the Trojaned shadow model (line 9). We repeat the process multiple times to generate a set of different Trojaned models. The sampling algorithm (line 3) and model training algorithm (line 9) will be different for different tasks.

Note that the Trojan distribution defined by Eqn.\ref{eqn:trigger-general} does not capture all kinds of Trojans. We will evaluate how the trained meta-classifier performs in detecting the unforeseen Trojaned models in Section \ref{sec:unseen-exp}. In addition, it is possible that some future Trojans may occur which are not in the format as Eqn.\ref{eqn:trigger-general}.
In that case, we can modify the jumbo distribution to include the new type of Trojans.
Therefore, we conclude that jumbo learning is a generic way to generate a variety of Trojaned models.

\vspace{-0.5em}
\begin{algorithm}[t]
\caption{The pipeline of jumbo learning to generate random Trojaned shadow models.}
\label{alg:jumbo}
\KwIn{Dataset $D=\{({\bf x}_i,y_i)\}_{i=1}^n$, number of Trojaned shadow models to train $m$.}
\KwOut{$models$: a set of $m$ random Trojaned shadow models.}
$models \leftarrow []$\;
\For{$u = 1, \ldots, m$}{
    ${\bf m}, {\bf t}, \alpha, y_t, p = \text{generate\_random\_setting}()$\;
    $D_{\emph{troj}} \leftarrow D$\;
    $indices = \text{CHOOSE}(n, int(n*p))$\;
    \For{$j~\text{in}~indices$}{
        ${\bf x}'_j, y'_j \leftarrow I({\bf x_j}, y_j; {\bf m}, {\bf t}, \alpha, y_t)$\;
        $D_{\emph{troj}} \leftarrow D_{\emph{troj}} ~\bigcup~({\bf x}'_j, y'_j)$\;
    }
    $f_u \leftarrow \text{train\_shadow\_model}(D_{\emph{troj}})$\;
    $models.\text{append}(f_u)$\;
}
return $models$
\end{algorithm}

\presubspace
\subsection{Meta-training}
\postsubspace
\label{sec:mna-troj-detect}



The defender will perform the meta-training algorithm based on the set of shadow models generated by jumbo learning.
Our meta-training consists of two goals: 1) find a feature extraction function to extract representation vectors of the shadow models and 2) train a meta-classifier to distinguish between benign and Trojaned shadow models. In the following, we will first introduce our designs of the feature extraction function and the meta-classifier. Then we introduce query-tuning, a meta-training algorithm to jointly optimize the two components. Finally we will talk about a baseline algorithm when the defender does not apply jumbo learning and only have a set of benign shadow models.

\paragraph{Feature Extraction Function Design}
We propose to feed a set of queries to the shadow model and use the output vector as its representation features. We have two intuitions for this design. First, Trojaned models will behave differently with benign models (i.e., have different distributions of outputs) on some query inputs. For example, their outputs differ a lot on inputs with Trojan triggers. Second, we can get the model output without accessing its internal structure. This allows us to detect Trojans in black-box scenarios.

Formally speaking, Let $\{(f_i, b_i)\}_{i=1}^m$ denote the shadow model dataset, where $f_i: \mathbb{R}^{d_x} \rightarrow \{1,\ldots,c\}$ is the shadow model and $b_i$ is a binary label indicating whether $f_i$ is Trojaned ($b_i=1$) or benign ($b_i=0$).
We will choose a set of $k$ query inputs $X=\{{\bf x}_1, \ldots, {\bf x}_k\}$ where ${\bf x}_i \in \mathbb{R}^{d_x}$ (we will discuss how these query inputs are chosen later). We will feed the queries into the shadow model $f_i$ and get $k$ output vectors $\{f_i({\bf x}_1), \ldots, f_i({\bf x}_k)\}$.
By concatenating all the output vectors, we can get a representation vector $\mathcal{R}_i(X)$ as the feature of the shadow model $f_i$:
\begin{equation}
    \label{eqn:mna-repr}
    \mathcal{F}(f_i) = \mathcal{R}_{i}(X) = [[f_i({\bf x}_1) || \ldots || f_i({\bf x}_k)]] \in \mathbb{R}^{ck}
\end{equation}
where $[[\cdot||\cdot||\cdot]]$ stands for the concatenation operation. We use $\mathcal{R}_i$ to denote $\mathcal{R}_i(X)$ if it does not lead to misunderstanding.

\paragraph{Meta-classifier Design}
Let $\emph{META}(\mathcal{R}_i;\theta) \in \mathbb{R}$ denote the meta-classifier where $\theta$ denotes the parameter of $\emph{META}$. We propose to use a two-layer fully connected neural network as the meta-classifier. The meta-classifier will take in the feature vector $\mathcal{R}_i$ and output a real-valued score indicating the likelihood of $f_i$ to be Trojaned.

\paragraph{Meta-Training Algorithm}
In meta-training, we would like to find the optimal values in the query set $X$ and meta-classifier parameters $\theta$. One simple solution is to randomly choose the query set $X$, pre-calculate all the representation vectors $\mathcal{R}_i$ and only optimize the meta-classifier. Having $\mathcal{R}_i$ and corresponding label $b_i$, the meta-training is simply to minimize the loss of a binary classifier via gradient-based optimization:
\begin{equation}
    \label{eqn:loss}
    \argmin_{ \theta } \sum_{i=1}^m L\bigg(\emph{META}(\mathcal{R}_i(X);\theta), b_i\bigg)
\end{equation}
where $L(\cdot, \cdot)$ is the loss function using binary cross entropy.
This approach achieves a not bad performance in Trojan detection.
However, the randomly sampled inputs may not help distinguish the benign and Trojaned models because Trojaned models behave similarly with benign models on most inputs.
Therefore, we can improve the performance by finding the optimal query inputs to provide the most useful information in the representation vectors.

To this end, we propose a \emph{query-tuning} technique to find the best query set for feature extraction, which is similar with the technique in Oh et al.~\cite{joon18iclr}.
The main idea is to jointly optimize the query set and the meta-classifier in order to minimize the training loss. The optimization goal thus becomes:
\begin{equation}
    \label{eqn:loss-tune}
    \argmin_{ \substack{\theta \\ X=\{{\bf x}_1, \ldots, {\bf x_k}\}}} \sum_{i=1}^m L\bigg(\emph{META}(\mathcal{R}_i(X);\theta), b_i\bigg)
\end{equation}
Note that the query set $\{{\bf x}_1, \ldots, {\bf x_k}\}$ does not appear explicitly in the optimization goal, but are included in the calculation of $\mathcal{R}_i(X)$. 

To optimize the goal in Eqn. \ref{eqn:loss-tune}, a key observation is that the entire calculation flow is differentiable: we first feed the query inputs into the shadow models, then use their output as the representation vectors of shadow models, and finally feed them into the meta-classifier. Since the shadow models and the meta-classifier are differentiable, we can directly calculate the gradient of the loss with respect to the input vectors $\{{\bf x}_1, \ldots, {\bf x}_k\}$. Thus, we can still apply the standard gradient-based optimization technique for solving Eqn. \ref{eqn:loss-tune}. In particular, we will first randomly sample each ${\bf x}_i$ from a Gaussian distribution. Then we iteratively update ${\bf x}_i$ and $\theta$ with respect to the goal in Eqn. \ref{eqn:loss-tune} to find the optimal query set. This workflow is illustrated in the right part of Figure~\ref{fig:workflow}.

Note that during the training process we need to access the internal parameters of the shadow models for calculating the gradient. However, this does not violate the black-box setting because the shadow models are trained by us and we can for sure access their parameters. During the inference process, we only need to query the black-box target model with the tuned inputs $\{{\bf x}_1, \ldots, {\bf x}_k\}$ and use the output for detection.

\paragraph{Baseline Meta-training algorithm without jumbo learning}
The previous meta-training algorithm requires a set of Trojaned shadow models generated by jumbo learning.
In the following, we will introduce a baseline meta-training algorithm which requires only benign shadow models and thus does not need jumbo learning.

We assume that the defender has only benign shadow models to train the meta-classifier. The standard way to train a machine learning model with only one-class data is novelty detection, where the model is trained to determine whether an input is similar with its training samples. As an example, one-class SVM~\cite{manevitz2001one} will train a hyper-plane which separates all the training data from the origin while maximizing the distance $\rho$ from the origin to the hyper-plane. An example of one-class SVM is shown in Appendix \ref{app:svm}.

In practice, we propose to use one-class neural network~\cite{chalapathy2018anomaly} which generalizes the one-class optimization goal to neural networks, so that our meta-classifier can still be the two-layer network structure. The optimization goal in one-class neural network with query-tuning is:
\begin{equation}
    \label{eqn:ocnn}
    \min_{\substack{\theta, \rho\\ X=\{{\bf x}_1, \ldots, {\bf x_k}\}}} \frac12\cdot l_2(\theta) + \frac1\nu\cdot\frac1m\sum_{i=1}^m \emph{ReLU}\big(\rho - \emph{META}(\mathcal{R}_i(X);\theta)\big) - \rho
\end{equation}
where $l_2(\theta)$ stands for the sum of Frobenius norm of all the parameters in the meta-classifier. We will use a gradient-based approach to do meta-training and find the optimal query set $X$ and model parameters $\theta$ which minimize the goal as in Eqn.~\ref{eqn:ocnn}.


\presubspace
\subsection{Target Model Detection}
\label{sec:target-model-detect}
\postsubspace

Let $X^*=\{{\bf x}_1^*,\ldots, {\bf x}_k^*\}$ and $\theta^*$ denote the optimal query set and parameters obtained by meta-training. Given a target model $f_{tgt}$, we can apply the optimized feature extraction function and meta-classifier to determine whether it is Trojaned or not. In particular, we first calculate its representation vector $\mathcal{R}_{tgt} = [[f_{tgt}({\bf x}_1^*) || \ldots || f_{tgt}({\bf x}_k^*)]]$. Then we can determine whether $f_{tgt}$ is Trojaned according to the meta-classifier's prediction $\emph{META}(\mathcal{R}_{tgt};\theta^*)$.
\newcommand{\mnist}{MNIST\xspace}
\newcommand{\cifar}{CIFAR10\xspace}
\newcommand{\speech}{SC\xspace}
\newcommand{\irish}{Irish\xspace}
\newcommand{\nlp}{MR\xspace}
\newcommand{\acc}{ACC\xspace}
\newcommand{\atk}{ASR\xspace}
\newcommand{\accs}{\acc-S\xspace}
\newcommand{\acct}{\acc-T\xspace}
\newcommand{\atks}{\atk-S\xspace}
\newcommand{\atkt}{\atk-T\xspace}

\newcommand{\mnistb}{\mnist-B\xspace}
\newcommand{\mnistp}{\mnist-P\xspace}
\newcommand{\mnistm}{\mnist-M\xspace}
\newcommand{\mnistl}{\mnist-L\xspace}
\newcommand{\cifarb}{\cifar-B\xspace}
\newcommand{\cifarp}{\cifar-P\xspace}
\newcommand{\cifarm}{\cifar-M\xspace}
\newcommand{\cifarl}{\cifar-L\xspace}
\newcommand{\irishm}{\irish-M\xspace}
\newcommand{\irishb}{\irish-B\xspace}
\newcommand{\speechm}{\speech-M\xspace}
\newcommand{\speechb}{\speech-B\xspace}
\newcommand{\nlpm}{\nlp-M\xspace}


\presecspace
\section{Experiment Setup}
\label{sec:setup}
\postsecspace

In this section, we will introduce the experiment setting. We first introduce the datasets we use, followed by the attack and defense setting in the experiments. Finally we introduce the setting of baselines which we compare with. Our code is publically available at \url{https://github.com/AI-secure/Meta-Nerual-Trojan-Detection}.

\presubspace
\subsection{Dataset}
\postsubspace

We conduct our evaluation on a variety of machine learning tasks, covering different types of datasets and neural networks.
For the vision tasks, we use the standard MNIST~\cite{lecun_cortes_burges} and CIFAR10~\cite{cifar} datasets. MNIST is a dataset of grayscale handwritten digits 0-9 and CIFAR10 is a dataset of RGB images of 10 classes. For the speech task we use the SpeechCommand dataset (SC) containing one-second audio command of 10 classes. For the tabular data we use the Smart Meter Electricity Trial data in Ireland dataset (\irish)~\cite{irish}. It consists of electricity consumption over weeks of two types of users (residential vs. commercial). For the natural languange data we use the Rotten Tomatoes movie review dataset (MR)~\cite{kim2014convolutional} which consists of movie reviews and the task is to determine whether a review is positive or negative.
We provide detailed introduction of each dataset and network structure in each task in Appendix~\ref{app:structure}.

As discussed in Section \ref{sec:def-cap}, we assume the defender only has a small set of clean data to help with the detection and the data is different from the model training set (which is owned by the attacker).
Therefore, for each dataset, we randomly sample 50\% of the training set as owned by the attacker and 2\% of the training data as the defender's clean dataset.

\presubspace
\subsection{Attack Settings}
\postsubspace
\label{sec:atk-def-setting}


Here we introduce the attack setting that are modelled in our jumbo distribution as in Section \ref{sec:jumbo}. Unforeseen attack strategies will be evaluated in Section \ref{sec:unseen-exp}, which include parameter and latent attack (since they will not poison the training dataset) and all-to-all attack (since the jumbo distribution only considers single-target attack).

For each dataset except \nlp, the attacker will generate 256 target models using modification attack and blending attack respectively. On the discrete \nlp dataset only modification attack applies. In the following we describe the attack setting for both approaches.

\paragraph{Trigger mask $\bf m$} For blending attack, the pattern size is the same as the input, so ${\bf m} = 1$ everywhere. For modification attack, trigger mask differs in different tasks. On \mnist and \cifar, we use a square pattern with random size from $2\times2$ to $5\times5$ at random location; on \speech, the pattern will be a consecutive part at random place, whose length is randomly sampled from $\{0.05,0.1,0.15,0.2\}$ seconds; on \irish, the pattern will be a consecutive part at random place, whose length is randomly sampled from $\{1,2,3,4,5\}$ hours; on \nlp, we will add a random phrase with 1 or 2 words at random place.

\paragraph{Trigger pattern $\bf t$} The pattern value will be generated in the same way for modification and blending attack. On \mnist and \cifar, each pixel value is uniformly sampled from $[0,1]$; on \speech and \irish, each signal value is uniformly sampled from $[0,0.2]$; on \nlp, each word is uniformly sampled from the vocabulary.

\paragraph{Transparency $\alpha$} There is no transparency for modification attack, so $\alpha=0$; for blending attack we uniformly sample $\alpha$ from $[0.8, 0.95]$.

\paragraph{Malicious label $y_t$} The malicious label for each Trojaned model is uniformly chosen from the output set of each task, e.g. from digit 0-9 for \mnist or from the 10 types of commands for \speech.

\paragraph{Data poisoning ratio $p$} The proportion of injected data in the data poisoning is uniformly sampled from $[0.05, 0.5]$ for all the tasks and attacks.

Besides the Trojaned model, we also train 256 benign target models using the attacker's dataset to evaluate the detection performance. These benign models are trained using the same setting except for different model parameter initialization.

\presubspace
\subsection{Defense Settings}
\postsubspace
In jumbo \Sys{}, the defender will generate 2048 Trojaned models using jumbo learning and 2048 benign models to train the meta-classifier. The defender will also generate 256 Trojaned and benign models for validation. In one-class \Sys{}, only benign models are trained.
The Trojaned models are generated in the same way as the attacker, except for the following difference:
\begin{enumerate}
    \item The models here are trained using the defender's dataset, whose size is much smaller than the attacker's dataset.
    \item The trigger shape will be either small or the same size as the input. There is 20\% probability that the trigger shape is the same as the input; otherwise it is sampled in the way as modification attack before.
    \item The transparency $\alpha$ will be sampled conditional on the trigger shape. If the trigger shape is the same as the input, then the $\alpha$ is uniformly sampled from $[0.8,0.95]$; otherwise, $\alpha$ will be 0 with 25\% probability and otherwise uniformly sampled from $[0.5,0.8]$.
\end{enumerate}
In addition, we ensure that the attack settings which already exist in the attacker's Trojan models will not be sampled again in the training of defender's shadow models.

We use the Adam optimizer~\cite{kingma2014adam} with learning rate 0.001 to train all the models, meta-classifiers and tune the queries. 
We choose the query number to be $k=10$ as it already works well in our experiment (i.e., we do not need to choose a larger number of queries). In practice, we find the performance is not sensitive to this choice.

Note that our pipeline does not apply directly to the \nlp task which has discrete input. We introduce the adaptation of our approach in this case in Appendix~\ref{sec:app-discrete}.

\presubspace
\subsection{Detection Baselines}
\postsubspace
In our evaluation, we compare with four existing works on Trojan attack detection as our baselines: Activation Clustering (AC), Neural Cleanse (NC), Spectral Signature (Spectral) and STRIP. We do not compare with DeepInspect~\cite{chen2019deepinspect} as the authors have not releases the code and their pipeline is rather complicated. We do not compare with SentiNet~\cite{chou2018sentinet} as it only works on image dataset and the time cost for model-level detection is high. We introduce the details in comparing these approaches in Appendix~\ref{sec:app-detect-baseline}.


\presecspace
\section{Experimental Evaluation}
\postsecspace
\label{sec:eval}
In this section, we present the results of using our pipelines to detect Trojaned models.

\presubspace
\subsection{Trojan Attacks Performance}
\postsubspace

We first show the classification accuracy and attack success rate of the shadow models and target models in Table \ref{table:model-atk}. Accuracy is evaluated on normal input and attack success rate is evaluated on Trojaned input. On \irish dataset, we use Area Under ROC Curve (AUC) as the metric instead of accuracy since this is a binary classification task on unbalanced dataset. The defender will not perform the modification and blending attacks, so we only show the accuracy of the benign shadow models.

We can see that the attacker successfully installs Trojans in the target models. The accuracy is similar with the benign target models, while achieving a high attack success rate in all the tasks. In addition, we also see an obvious accuracy gap between benign shadow models and target models.
This matches our assumption that the model consumer cannot train a high-quality model based on his small dataset.
Therefore, he needs to use the shared model instead of training a model using his own clean dataset. 

\begin{table}[tbp]
    \centering
    \caption{The classification accuracy and attack success rate for the shadow and target models. -M stands for modification attack and -B stands for blending attack.}
    \footnotesize
    \begin{tabular}{l|c|c|c|c}
       \hline 
       \multirow{3}{*}{\textbf{Models}} & \multicolumn{2}{c|}{\textbf{Shadow Model}} & \multicolumn{2}{c}{\textbf{Target Model}} \\
       \cline{2-5}
        & \textbf{Accuracy} &  \makecell{\textbf{Success}\\\textbf{Rate}} & \textbf{Accuracy} & \makecell{\textbf{Success}\\\textbf{Rate}} \\
        \hline 
        \hline 
        \mnist & 95.14\% & - & 98.47\% & - \\
        \hline 
        \mnistm & - & - & 98.35\% & 99.76\% \\
        \hline 
        \mnistb & - & - & 98.24\% & 99.68\% \\
        \hline 
        \hline 
        \cifar & 39.31\% & - & 61.34\% & - \\
        \hline 
        \cifarm & - & - & 61.23\% & 99.65\% \\
        \hline 
        \cifarb & - & - & 59.52\% & 89.92\% \\
        \hline 
        \hline 
        \speech & 66.00\% & - & 83.43\% & - \\
        \hline 
        \speechm & - & - & 83.20\% & 98.66\% \\ 
        \hline 
        \speechb & - & - & 83.56\% & 98.82\% \\ 
        \hline 
        \hline 
        \irish & 79.71\% & - & 95.88\% & - \\
        \hline 
        \irishm & - & - & 94.17\% & 95.78\% \\
        \hline 
        \irishb & - & - & 93.62\% & 92.79\%\\ 
        \hline 
        \hline 
        \nlp & 72.61\% & - & 74.69\% & - \\
        \hline 
        \nlpm & - & - & 74.48\% & 97.47\% \\ 
        \hline 
    \end{tabular}

    \label{table:model-atk}
    \vspace{-1em}
\end{table}

\presubspace
\subsection{Detection Performance}
\postsubspace
\label{sec:perf}
\begin{table*}[!th]
    \centering
    \scriptsize
	\caption{The detection AUC of each approach. -M stands for modification attack and -B stands for blending attack.}
    \begin{tabular}{c|c|c|c|c|c|c|c|c|c}
        \hline
        \textbf{Approach} & \textbf{\mnistm} & \textbf{\mnistb} & \textbf{\cifarm} & \textbf{\cifarb} & \textbf{\speechm} & \textbf{\speechb} & \textbf{\irishm} & \textbf{\irishb} & \textbf{\nlpm}\\
        \hline
        \hline
        AC~\cite{chen2018detecting} & 73.27\% & 78.61\% & 85.99\% & 74.62\% & 79.69\% & 82.86\% & 56.14\% & 93.48\% & 88.26\% \\
        \hline
        NC~\cite{wang2019neural} & 92.43\% & 89.94\% & 53.71\% & 57.23\% & 91.21\% & 96.68\% & \xmark & \xmark & \xmark \\
        \hline
        Spectral~\cite{tran2018spectral} & 56.08\% & $\leq$50\% & 88.37\% & 58.64\% & $\leq$50\% & $\leq$50\% & 56.50\% & $\leq$50\% & \bf 95.70\%\\
        \hline
        STRIP~\cite{gao2019strip} & 85.06\% & 66.11\% & 85.55\% & 81.45\% & 89.84\% & 85.94\% & $\leq$50\% & $\leq$50\% & \xmark \\
        \hline
        \hline
        \makecell{\Sys{} (One-class)} & 61.63\% & $\leq$50\% & 63.99\% & 73.77\% & 87.45\% & 85.91\% & 94.36\% & \bf 99.98\% & $\leq$50\%\\
        \hline
        \makecell{\Sys{} (Jumbo)} & \bf 99.77\% & \bf 99.99\% & \bf 91.95\% & \bf 95.45\% & \bf 99.90\% & \bf 99.83\% & \bf 98.10\% & \bf 99.98\% & 89.23\%\\
        \hline
    \end{tabular}
    \label{tab:result-cv}
\end{table*}


Using the setup in \refsect{setup}, we compare our jumbo \Sys{} approach and one-class \Sys{} approach with the four baseline approaches.
We use the AUC as the metric to evaluate the detection performance. The results are shown in Table \ref{tab:result-cv}. We use \xmark~to show that the approach cannot be applied on the experiment setting.

As the discussion in Section 
\ref{sec:backg-baseline} goes, all the baseline approaches have some assumptions on the attacks, so they only work on a few tasks and cannot keep high performance through all the tasks.
On the other hand, we would like to point out that Spectral and STRIP are not aimed to perform model-level Trojan detection and we design the pipeline to adjust them to detect Trojaned models (i.e., to average score for each data in the training set). Therefore, it is unfair to compare our results with theirs directly and claim that their works do not work well, but it does show that no existing work can achieve a good performance on the task of model-level Trojan detection.

As a comparison, our Jumbo \Sys{} approach achieves over 90\% detection AUC in all but one of the experiments that cover different datasets and attacks and the average detection AUC reaches over 97\%. 
In addition, this approach outperforms all the baseline approaches except for the NLP task (89.23\% vs. 95.70\% of Spectral). However, Jumbo \Sys{} does not need to access the training dataset as Spectral does and only queries the embedding layer; we consider the results comparable with that of the baselines. 
On the other hand, our one-class approach is good on some tasks but fails on others. On some tasks it is even worse than random guesses.
We leave the interpretation of this interesting phenomenon as our future work.
We include the ROC curve of the detection performance as well as the isolation experiments of query tuning in Appendix~\ref{sec:app-other-exp}.

\presubspace
\subsection{Impact of Number of Shadow Models}
\begin{figure}[tbp]
\vspace{-1em}
    \centering
    \includegraphics[width=0.23\textwidth]{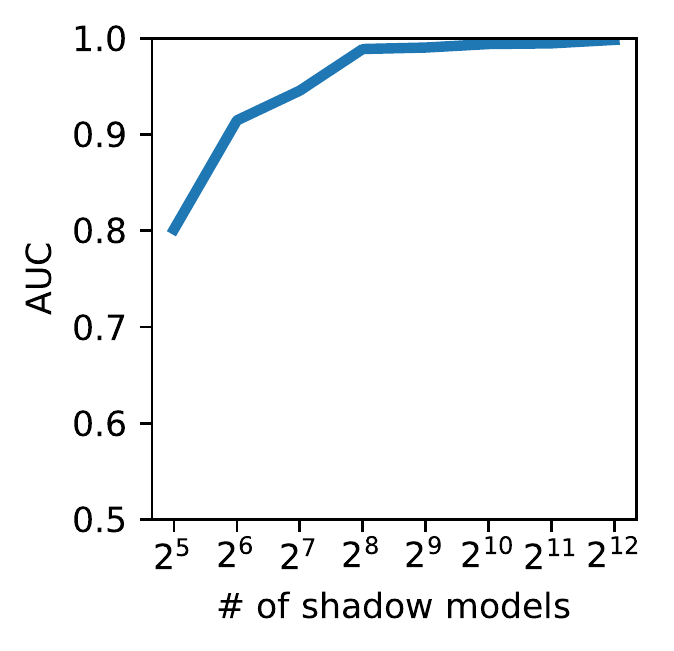}
    \includegraphics[width=0.23\textwidth]{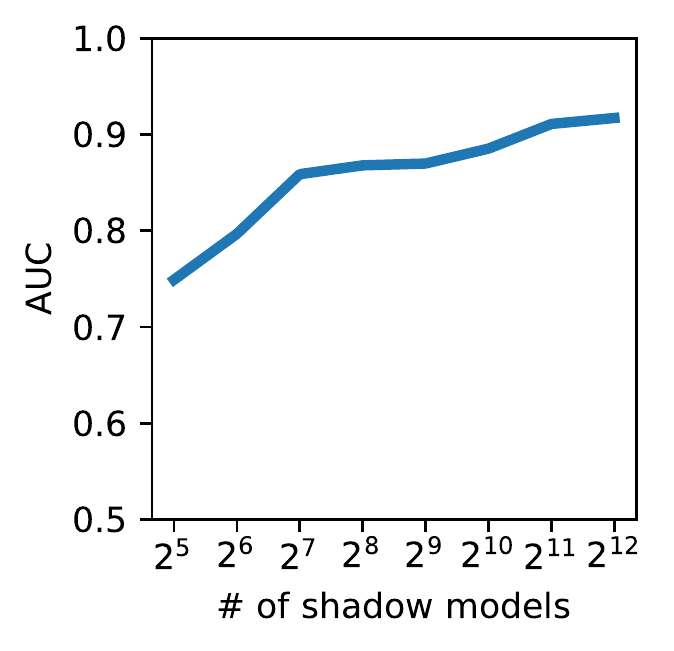}
    \caption{Detection AUC with respect to the number of shadow models used to train the meta-classifier on \mnistm (left) and \cifarm(right).}
    \label{fig:num-shadow}
\vspace{-1em}
\end{figure}

\postsubspace
\label{sec:number-of-shadow}
\label{sec:num-shadow}
In Figure \ref{fig:num-shadow}, we demonstrate the impact of using different number of shadow models in training the meta-classifier on the \mnistm and \cifarm tasks. Our approach can achieve a good result even with a small number of shadow models (e.g., only 128 benign models + 128 Trojaned models). With more shadow models, the accuracy continues to grow. 
Defenders with different computational resources can make a trade-off between the number of shadow models and the detection performance based on their needs.
We include more discussion on the efficiency of our approach in Section \ref{sec:running-time}.

\presubspace
\subsection{Running Time Performance}
\label{sec:running-time}
\postsubspace

\begin{table}[tbp]
    \footnotesize
    \centering
    \caption{Running time required to detect one target model on \mnistm.}
    \begin{tabular}{c|c}
        \hline
        Approach & Time (sec)\\
        \hline
        AC  & 27.13 \\
        NC  & 57.21\\
        Spectral  & 42.55\\
        STRIP  & 738.5\\
        \Sys{}  & $\mathbf{2.629 \times 10^{-3}}$\\
        \Sys{} (offline preparation time) & $\sim 4096 \times 12 + 125$\\
        \hline
    \end{tabular}
    \label{tab:time}
    \vspace{-1.5em}
\end{table}

We compare the detection running time of each approach on the \mnistm task in Table \ref{tab:time}. The experiment is run on one NVIDIA GeForce RTX 2080 Graphics Card. The running time of our pipeline contains two parts - the offline training part which includes shadow model generation and meta-classifier training, and the inference part that detects Trojaned target models. It takes around 12 seconds to train each shadow model and 125 seconds to train the meta-classifier. Therefore, the offline part needs $4096\times12+125$ seconds, which is around 14 hours. On the other hand, in the inference part we only need to query the target model with our tuned queries and apply the meta-classifier, which is very efficient and takes only 2.63ms.

As a comparison, we see that the running time to detect Trojans using baseline approaches varies between 27 seconds to 738 seconds. 
We would like to emphasize that we only need to perform the offline part once for each task. That is, as long as we have trained the meta-classifier on \mnist, we can apply it to detect any Trojaned \mnist model. It takes only several milliseconds to apply the trained \Sys{} on one target model. In contrast, other approaches have to re-run their entire algorithm whenever they are provided with a new model. Therefore, our approach is more efficient when the defender needs to detect Trojans on a number of target models on the same task. Moreover, as demonstrated in \refsect{number-of-shadow}, the model consumer could also generate a smaller number of shadow models to make a trade-off between computation overhead and Trojan detection performance.

\begin{table*}[t]
    \centering
    \caption{Examples of unforeseen Trojan trigger patterns and the detection AUC of jumbo MNTD on these Trojans.}
    \label{tab:result-unseen-pattern-general}
    \begin{tabular}{c|c|c|c|c|c|c}
        \hline
        Trojan & \multicolumn{3}{c|}{MNIST} & \multicolumn{3}{c}{CIFAR-10} \\
        \cline{2-7}
        Shape & Pattern Mask & Trojaned Example & Detection AUC & Pattern mask & Trojaned Example & Detection AUC\\
        \hline
        \raisebox{1.5em}{Apple} & \includegraphics[width=0.1\textwidth]{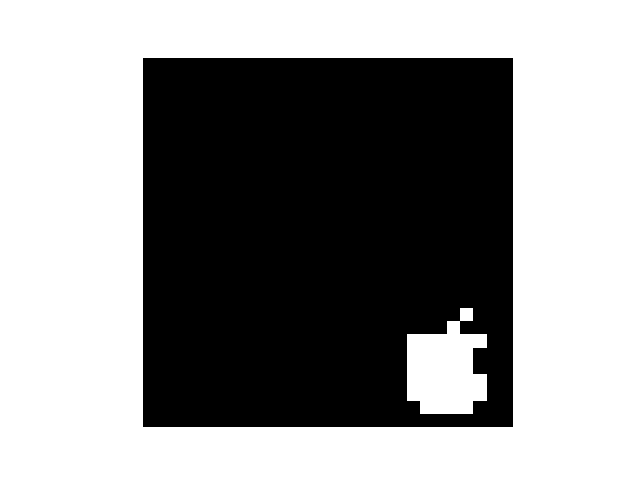} & \includegraphics[width=0.1\textwidth]{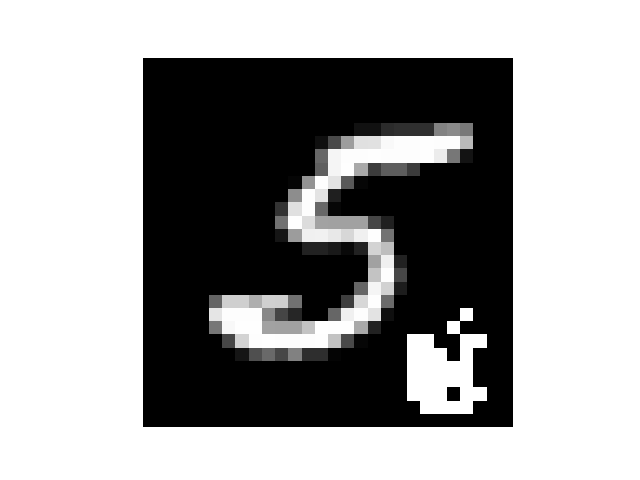} & \raisebox{1.5em}{96.73\%} & \includegraphics[width=0.1\textwidth]{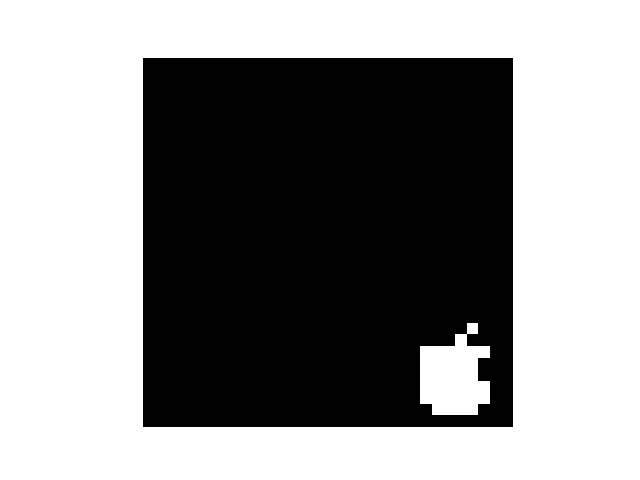} & \includegraphics[width=0.1\textwidth]{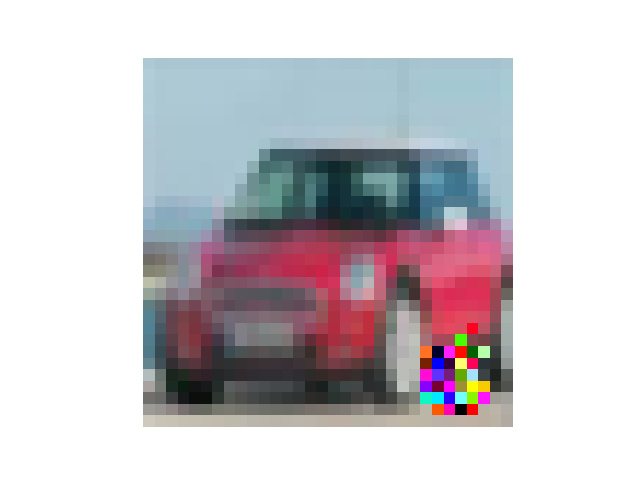} & \raisebox{1.5em}{89.38\%} \\
        \hline
        
        \raisebox{1.5em}{Corners} & \includegraphics[width=0.1\textwidth]{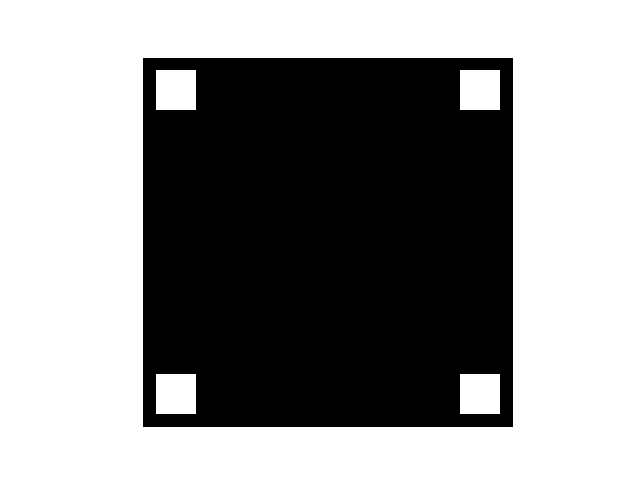} & \includegraphics[width=0.1\textwidth]{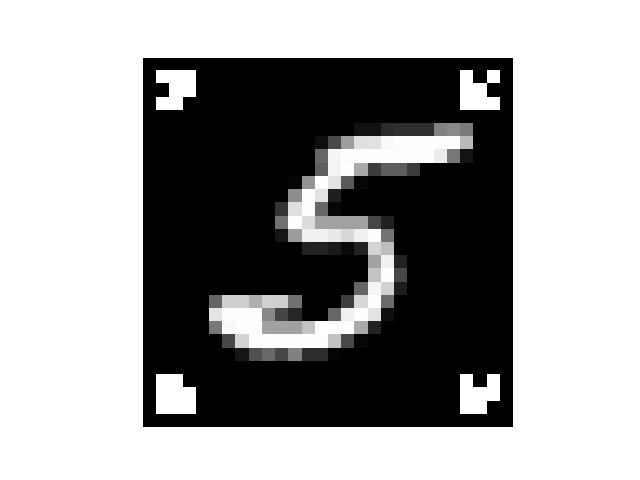} & \raisebox{1.5em}{98.74\%} & \includegraphics[width=0.1\textwidth]{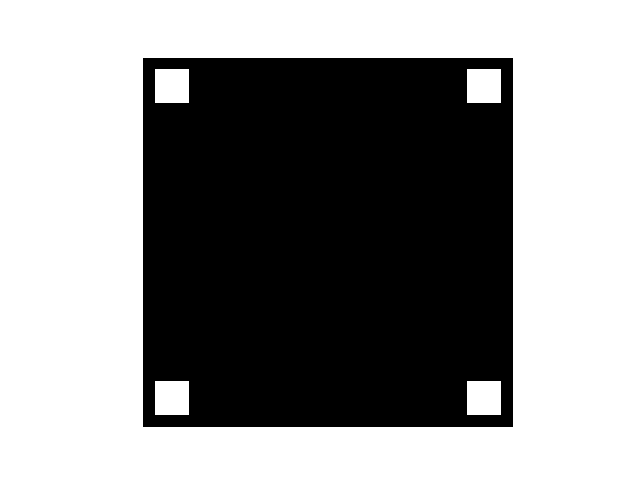} & \includegraphics[width=0.1\textwidth]{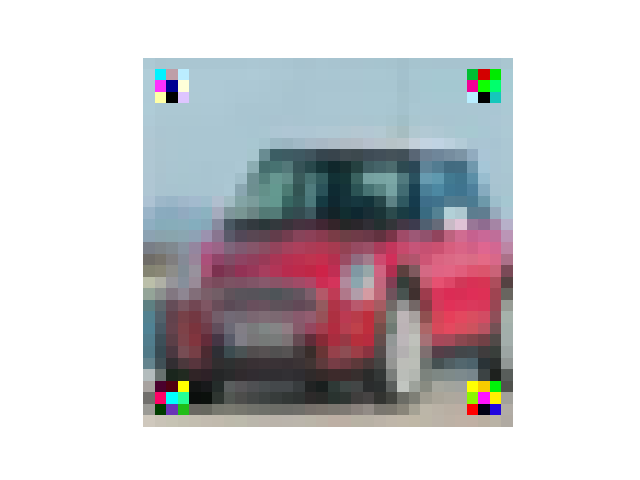} & \raisebox{1.5em}{93.09\%} \\
        \hline
        
        \raisebox{1.5em}{Diagonal} & \includegraphics[width=0.1\textwidth]{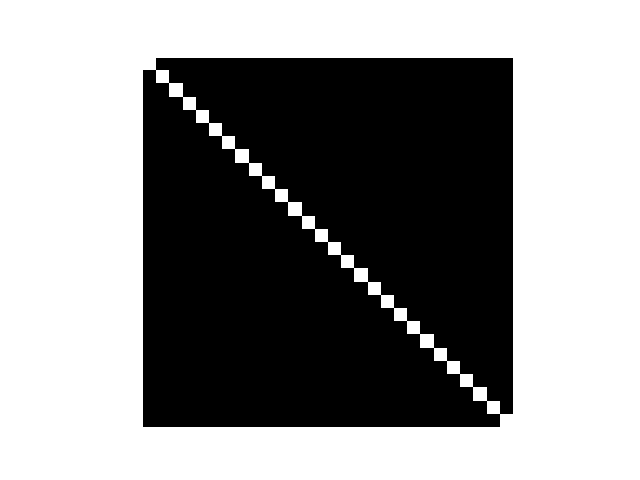} & \includegraphics[width=0.1\textwidth]{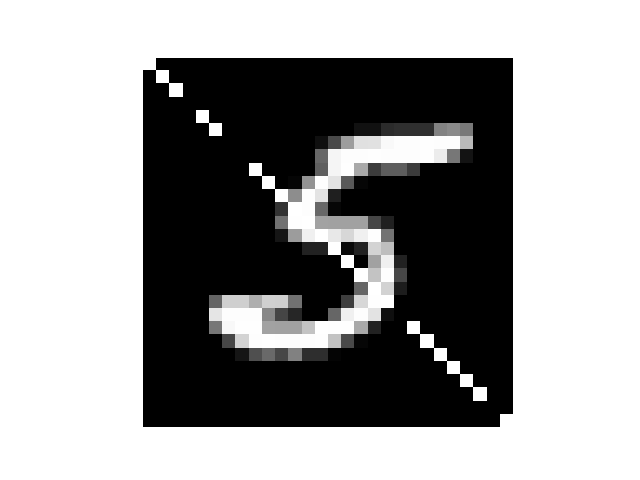} & \raisebox{1.5em}{99.80\%} & \includegraphics[width=0.1\textwidth]{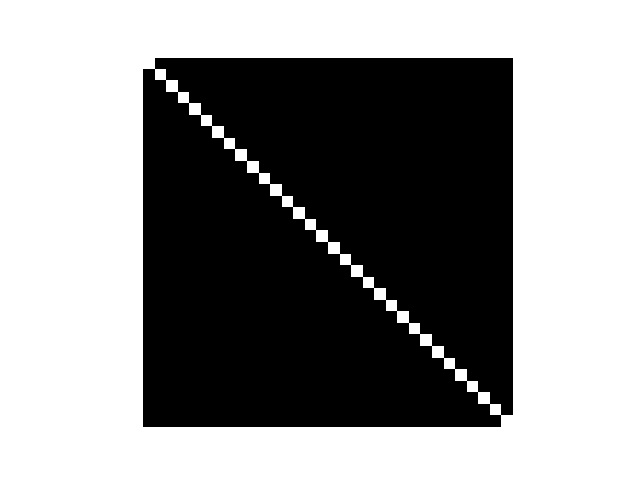} & \includegraphics[width=0.1\textwidth]{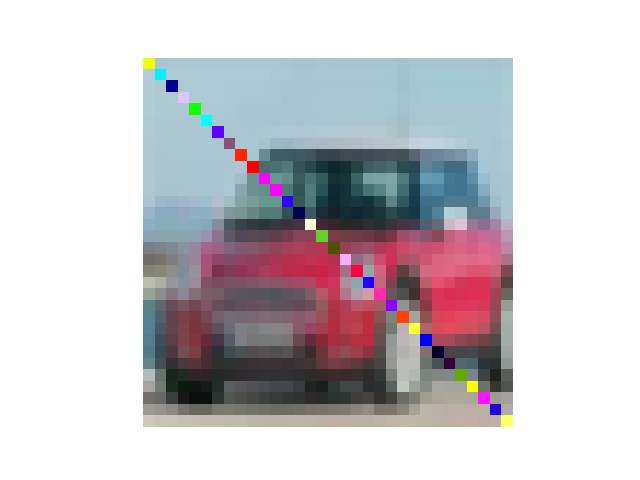} & \raisebox{1.5em}{97.57\%} \\
        \hline
        
        \raisebox{1.5em}{Heart} & \includegraphics[width=0.1\textwidth]{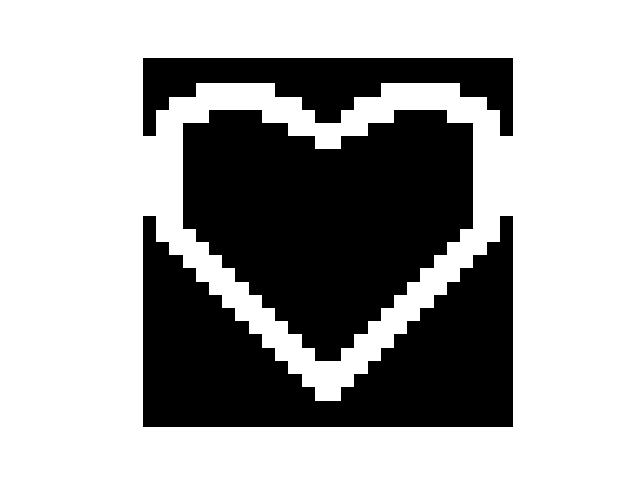} & \includegraphics[width=0.1\textwidth]{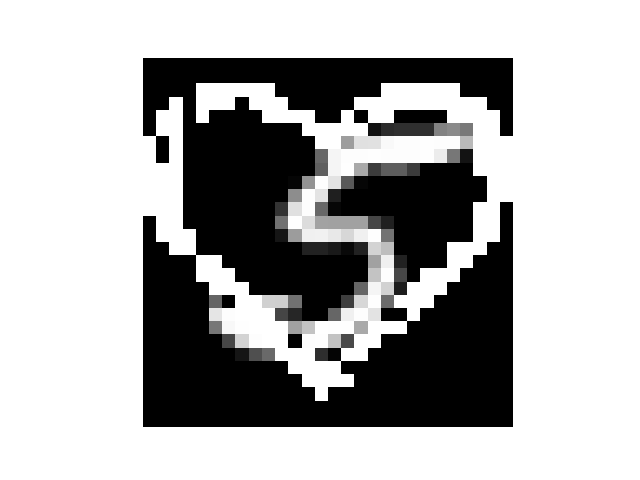} & \raisebox{1.5em}{99.01\%} & \includegraphics[width=0.1\textwidth]{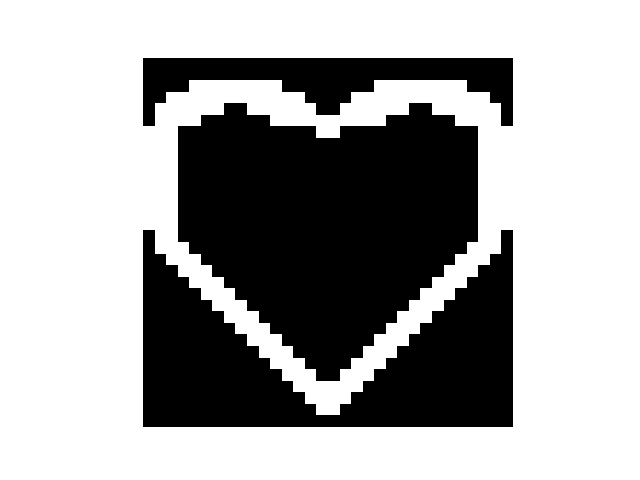} & \includegraphics[width=0.1\textwidth]{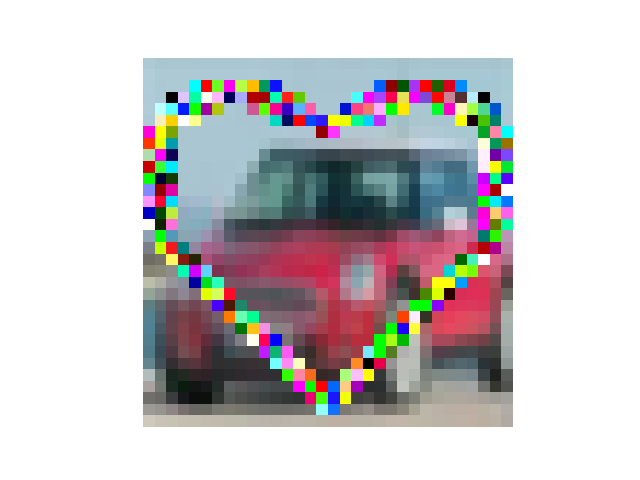} & \raisebox{1.5em}{93.82\%} \\
        \hline
        
        \raisebox{1.5em}{Watermark} & \includegraphics[width=0.1\textwidth]{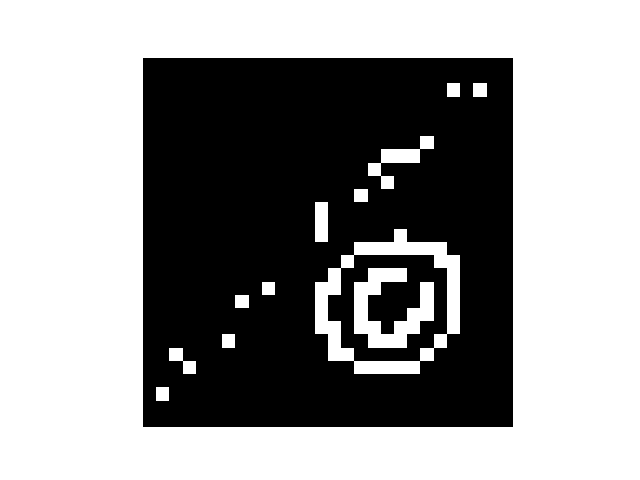} & \includegraphics[width=0.1\textwidth]{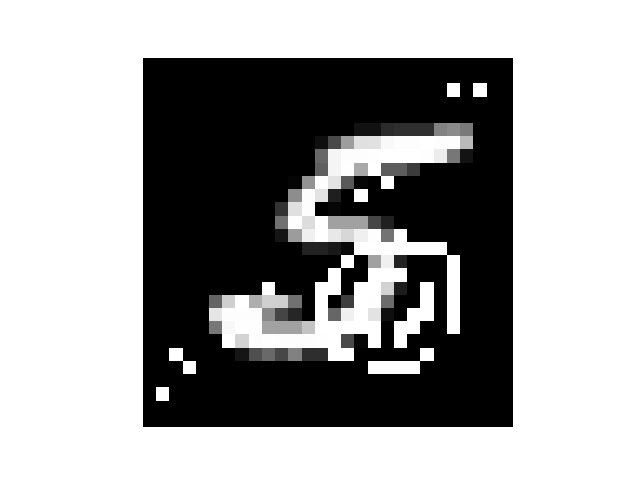} & \raisebox{1.5em}{99.93\%} & \includegraphics[width=0.1\textwidth]{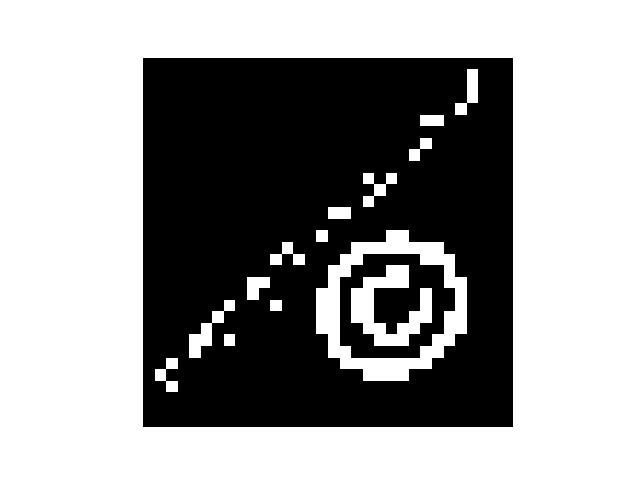} & \includegraphics[width=0.1\textwidth]{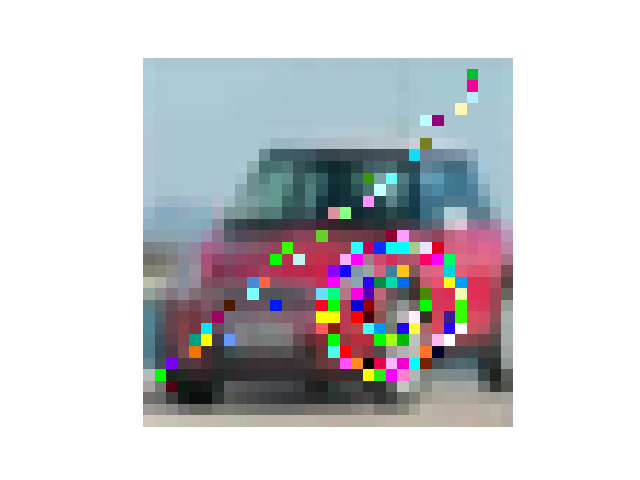} & \raisebox{1.5em}{97.32\%} \\
        \hline
    \end{tabular}
\end{table*}

\presecspace
\section{Generalization on Unforeseen Trojans}
\label{sec:unseen-exp}
\postsecspace

In this section, we will evaluate jumbo \Sys{} using Trojaned models that are not modelled by the jumbo distribution in the training process. Unless specified, we will train 256 Trojaned models for each attack setting. We empirically verified that all the attack settings are successful to install Trojans in the model.

\presubspace
\subsection{Generalization on Trigger Patterns}
\postsubspace

We first evaluate the meta-classifier using unforeseen trigger patterns.
We collect some Trojan shapes on vision tasks used in previous works~\cite{trojannn,gao2019strip} as well as some newly designed Trojan shapes, as shown in Table~\ref{tab:result-unseen-pattern-general}.
A significant difference between these shapes and those in jumbo learning is that they will change a large number of pixels;
in contrast, the patterns in jumbo learning change at most $5\times5$ pixels.
For each pattern type, we use the same mask but with randomly generated pixel values. We train the Trojaned models with these patterns and apply the jumbo MNTD pipeline to detect them.

The detection results are shown in Table~\ref{tab:result-unseen-pattern-general}. We can see that the trained meta-classifier achieves similar detection results as before. This shows that our detection approach generalizes well to a variety of trigger patterns even if they are not considered in the training process. We provide further experiment results on the generalization to unforeseen trigger patterns in Appendix~\ref{sec:app-other-exp}.

\presubspace
\subsection{Generalization on Malicious Goals}
\postsubspace
All the models in the jumbo distribution aims at the single target attack, i.e., change the label of a Trojaned input to be a specific class. Here, we consider another type of malicious behavior, all-to-all attack. In particular, for a $c$-way classification model, the label of a Trojaned input which originally belongs to the $i$-th class will be changed to the $((i+1)~\text{mod}~c)$-th class. We will evaluate whether our meta-classifier can detect Trojaned models with all-to-all attack.
For MNIST, we adopt the same setting as in \cite{chen2018detecting} and add a four-pixel pattern at the right bottom corner. For other tasks, we use the same attack setting as the modification attacks on them and change the attack goal to be all-to-all attack.
We empirically find that the all-to-all attack cannot work on \speech and \nlp (the attack success rate is low), so we do not include them as in the detection task.

The results are shown in Table \ref{tab:result-unseen-behav}. We see that NC and STRIP cannot perform well in detecting these kinds of Trojans, as we have discussed in Section \ref{sec:backg-baseline}.
Our approach still achieves a good performance in detecting this unforeseen type of malicious goal, reaching over 98\% detection AUC for all the three tasks. AC outperforms us on MNIST by 0.05\% while we outperform them by 20\% and 10\% on the other two tasks. In addition, we do not require access to the dataset as AC does. These results show that our detection pipeline requires a weak assumption and is general to different tasks.

\begin{table}[t]
    \centering
    \caption{The detection AUC of each approach against all-to-all Attack.}
     \footnotesize
    \label{tab:result-unseen-behav}
    \begin{tabular}{c|c|c|c}
        \hline
         Approach & \mnist & \cifar & \irish \\
        \hline
         AC & \bf 100.00\% & 77.41\% & 90.94\% \\
         NC & 51.46\% & 52.34\% & \xmark \\
         Spectral & 84.36\% & $\leq$50\% & 68.02\% \\
         STRIP & 62.60\% & $\leq$50\% & $\leq$50\% \\
         \Sys{} (One-class) & 97.09\% & 70.38\% & 99.98\% \\
         \Sys{} (Jumbo) & 99.95\% & \bf 98.62\% & \bf 100.00\% \\
        \hline
    \end{tabular}
\vspace{-1em}
\end{table}

\presubspace
\subsection{Generalization on Attack Approaches}
\postsubspace

In the jumbo distribution we use poisoning attack and change the training dataset to generate Trojaned models. However, we introduced four types of attacks in Section \ref{sec:attacks} of which only modification attack and blending attack will insert Trojan by poisoning the dataset. In this section, we will evaluate how the meta-classifier performs in detecting the other two kinds of unforeseen attack approaches.

We evaluate the two attacks on vision tasks since we empirically find that the attack success rate is usually low on other tasks. For parameter attack, we add a $4\times4$ pattern for \mnist and $8\times8$ for \cifar. For latent attack, we follow the same setting in ~\cite{yao2019latent} and add a $5\times5$ pattern for both tasks.
The shadow and target models in latent attack are fine-tuned to the user's task before we perform our detection.

The results are shown in Table \ref{tab:result-unseen-atk}. Note that the attack will not poison the training dataset, so AC, Spectral and STRIP cannot be applied to detect such Trojans. We see that our model can detect these Trojaned models well. In addition, we emphasize that the latent attack appears after we first proposed our pipeline and we did not tailor our method in order to detect it. This shows that our approach has good generalizability in detecting unforeseen Trojan attack approaches.

\begin{table}[t]
    \centering
    \caption{The detection AUC of \Sys{} and neural cleanse on parameter attack (denoted by -P) and latent attack (denoted by -L). Other input-level and dataset-level detection techniques are not included as they cannot be applied in detecting these attacks.}
     \footnotesize
    \label{tab:result-unseen-atk}
    \begin{tabular}{c|c|c|c|c}
        \hline
         Approach & \mnistp & \mnistl & \cifarp & \cifarl \\
        \hline
         NC & $\leq$50\% & 95.02\% & 53.12\% & 83.79\%\\
         \Sys{}(one-class) & $\leq$50\% & 98.83\% & $\leq$50\% & $\leq$50\% \\
         \Sys{}(Jumbo) & \bf 99.99\% & \bf 99.07\% & \bf 98.87\% & \bf 92.78\%\\
        \hline
    \end{tabular}
\vspace{-1em}
\end{table}

\presubspace
\subsection{Generalization on Model Structures}
\label{sec:transfer}
\postsubspace

In \refsect{perf}, we evaluate \Sys with the assumption that the defender knows the target model architecture.
However, in some cases the defender may not have such knowledge. This problem might be solved by existing techniques which infer the structure of a black-box model~\cite{joon18iclr}. Nevertheless, we would like to evaluate how \Sys{} performs when generalizing to unforeseen model structures.

We perform our evaluation on the more complicated dataset ImageNet~\cite{imagenet_cvpr09}, as many different structures have been proposed to achieve a good performance on ImageNet. In particular, we use its subset on dog-vs-cat, which is a binary classification task between dogs and cats, containing 20,000 training cases and 5,000 testing cases. Here we assume that the defender owns 10\% of the training set instead of 2\% in previous assumption, since the ImageNet models are more difficult to train; the attacker still owns 50\%.

To evaluate the generalization to unforeseen model structures, we use six different structures in the model pool: (1) ResNet-18~\cite{he2016deep}, (2) ResNet-50~\cite{he2016deep}, (3) DenseNet-121~\cite{huang2017densely}, (4) DenseNet-169~\cite{huang2017densely}, (5) MobileNet v2~\cite{sandler2018mobilenetv2} and (6) GoogLeNet~\cite{szegedy2015going}. 
We use one of the six structures as the target model at a time.
The attacker will generate 32 target models of the structure, 16 benign and 16 Trojaned using jumbo distribution.
The defender will train 64 shadow models (32 benign and 32 jumbo Trojaned) for each of the other five structures, which generates 640 shadow models in total. He will use these shadow models to train the meta-classifier and detect whether the target models contain Trojan or not. Thus, the target model structure will never be seen in the training of the meta-classifier. The experiment is repeated for each of the six structures.

The results are reported in Table~\ref{tab:result_transfer_imagenet}. We see that all the AUCs are higher than 80\%, showing a good transferability even on complicated tasks like ImageNet.
Note that here we only use 64 models for each structure to train the meta-classifier due to time efficiency.
According to Section \ref{sec:number-of-shadow}, the results in Table~\ref{tab:result_transfer_imagenet} could be further improved by training more shadow models. The results demonstrate that \Sys{} is applicable to complicated tasks and generalizes well to unforeseen model structures.

\begin{table}[t]
    \centering
    \caption{The detection AUC of \Sys{} on unforeseen model structures on ImageNet Dog-vs-Cat. The meta-classifier for each model structure are trained using all models except the ones in target model structure.}
     \footnotesize
    \label{tab:result_transfer_imagenet}
    \begin{tabular}{c|c|c}
        \hline
         ResNet-18 & ResNet-50 & DenseNet-121 \\
        \hline
         81.25\% & 83.98\% & 89.84\%  \\
        \hline
         \hline
        DenseNet-169 & MobileNet & GoogLeNet \\
        \hline
         82.03\% & 87.89\% & 85.94\% \\
        \hline
    \end{tabular}
\vspace{-1em}
\end{table}

\presubspace
\subsection{Generalization on Data Distribution}

In previous studies, we assume that the data collected by the defender follows the same distribution as the model's training data. In this section, we study the case where the defender use alternative data which is similar but not the same distribution. We consider two alternatives for MNIST and CIFAR dataset - USPS digit dataset~\cite{hull1994database} and TinyImageNet dataset~\cite{imagenet_cvpr09}. The USPS dataset includes $16\times 16$ grayscale images of digit 0-9. It contains 7291 train and 2007 test images. We will reshape the images into $28\times 28$ so that it is same as MNIST. The TinyImageNet contains $64\times 64$ images of 200 classes where each class has 500 training, 50 validation and 50 test images. We hand-picked 10 classes to correspond to the labels in CIFAR-10.
The chosen classes are shown in Appendix~\ref{sec:app-label-map}.
We reshape the images into $32 \times 32$ to be the same as CIFAR-10.

In the experiments, we will train the shadow models using the USPS and TinyImageNet dataset instead of the 2\% of MNIST and CIFAR-10 dataset. The models trained on these alternative datasets can achieve 81.63\% accuracy on MNIST and 33.97\% on CIFAR-10. Then we train a meta-classifier and evaluate them on the target models of \mnistm, \mnistb, \cifarm, \cifarb as in Table \ref{tab:result-cv}. We find that the meta-classifier using USPS achieves 98.82\% detection AUC on \mnistm and 99.57\% on \mnistb; meta-classifier using TinyImageNet achieves 83.41\% AUC on \cifarm and 93.78\% on \cifarb. We can see that the the meta-classifier still achieves good detection performance, though it is slightly worse compared with the case when we use the same data distribution. This shows that the defender can use an alternative dataset to train the shadow models.

\postsubspace
\presecspace
\section{Adaptive Attack and Countermeasure}
\postsecspace
\begin{table*}[!th]
    \centering
    \scriptsize
	\caption{The detection AUC of \Sys{}-robust and its detection performance against strong adaptive attack.}
    \begin{tabular}{c|c|c|c|c|c|c|c|c|c}
        \hline
        \textbf{Approach} & \textbf{\mnistm} & \textbf{\mnistb} & \textbf{\cifarm} & \textbf{\cifarb} & \textbf{\speechm} & \textbf{\speechb} & \textbf{\irishm} & \textbf{\irishb} & \textbf{\nlpm}\\
        \hline
        \hline
        \makecell{\Sys{}-robust} & 99.37\% & 99.54\% & 96.97\% & 84.39\% & 96.61\% & 91.88\% & 99.92\% & 99.97\% & 96.81\%\\
        \hline
        \makecell{\Sys{}-robust\\(under attack)} & 88.54\% & 81.86\% & 94.83\% & 75.60\% & 88.86\% & 90.45\% & 97.27\% & 88.79\% & 94.78\% \\
        \hline
    \end{tabular}
    \label{tab:result-robust}
    \vspace{-1.5em}
\end{table*}

In this section, we consider a strong attacker who adapts their approach to evade MNTD and then 
extend our technique to be robust to such attacks.

\presubspace
\subsection{Strong Adaptive Attack}
\label{sec:adapt-atk}
\postsubspace
We consider a strong attacker who wishes to evade \Sys{}. We assume that the adversary has full knowledge of the detection pipeline, 
including the specific parameters of the meta-classifier $\emph{META}$ \emph{and} the tuned query input set $\{ {\bf x}_1, \ldots, {\bf x}_k\}$.
The goal of the attacker is to construct a Trojaned model that will be classified as benign by our \Sys{}.

With full knowledge of the \Sys{} system, the attacker can evade the detection by incorporating the prediction of \Sys{} in its training process. Suppose the original training loss for a Trojaned model $f$ is $L_{train}$ and the goal is:
\begin{align}
    \min_{f}~ L_{train}(f)
\end{align}
For example, on classification tasks, $L_{train}$ is the mean cross entropy loss between model predictions and ground truth labels over all the benign and Trojaned data. We denote $L_{mal}$ to be output of the meta-classifier on $f$. Recall that:
\begin{align}
    L_{mal}(f) &= \emph{META}(\mathcal{F}(f);\theta)\\
    \mathcal{F}(f) &= [[ f({\bf x}_1) || \ldots || f({\bf x}_k) ]]
\end{align}
Large $L_{mal}$ indicates that the model is evaluated as Trojaned by the \Sys{} system. Since the attacker has full knowledge of the detection system, he can calculate $L_{mal}$ during training and aims to keep it small. In particular, the attacker can explicitly add $L_{mal}$ into the training process and change the training goal as:
\begin{align}
    \label{eqn:adapt}
    \min_{f}~ L_{train} + \lambda \cdot L_{mal}
\end{align}
where $\lambda$ is a chosen parameter balancing model performance and evasion success rate.
With full knowledge of the \Sys{} system, the attacker can perform back-propagation to optimize the loss function directly. 
In practice, we use $\lambda = 1.0$ which works well for the adaptive attacks and we find the result not sensitive to this choice. In particular, the Trojaned model can always evade the detection of \Sys{} using the attack while incurring only negligible decrease in model accuracy (i.e., utility) and attack success rate.

\presubspace
\subsection{Countermeasure - \Sys{}-robust}
\postsubspace
The key point in the strong adaptive attack is that the adversary has full access to the meta-classifier parameters and query inputs. Hence, he can intentionally optimize the Trojaned model to make it look benign to the meta-classifier. In practice, the defender can avoid it by keeping the model parameters as a secret. Nevertheless, we will propose a robust version of our system, \Sys{}-robust, to counteract the strong attacker with full knowledge of our system.

The core idea of \Sys{}-robust is that during test time we will set part of the system parameters to be random values. Hence, the attacker cannot know what the parameters are and thus cannot calculate the exact value of $L_{mal}$. In particular, in \Sys{}-robust we will: 1) sample a random meta-classifier by setting its parameters values to be random numbers sampled from the normal distribution; 2) use our training set of shadow models to tune the queries only, while keeping the random meta-classifier unchanged; 3) we will use the tuned inputs along with the random meta-classifier to analyze a model and classify it as benign or Trojaned.

To guarantee that the attacker does not know the random parameters of the meta-classifier, the defender can re-sample them for each detection task. This would increase the detection cost since the defender needs to retrain the meta-classifier whenever the parameters are sampled. But as discussed in Section~\ref{sec:running-time}, the expensive part of \Sys{} is to train the shadow models, which needs only be done once; training the meta-classifier is comparatively fast. Additionally, a random meta-classifier could be reused for verifying an entire batch of models to be classified as Trojaned or benign; as long as the adversary does not know the random parameters, the defense remains robust.

From the attacker's side, he still wants to generate Trojaned models which can evade the detection with full knowledge of \Sys{}-robust. However, this time he does not know the random parameters of the meta-classifier and the corresponding query set. Hence, we assume that the attacker will first sample the random parameters of the meta-classifier on his own and tune the queries. Then he can apply the same technique as in Section \ref{sec:adapt-atk} to generate the Trojaned model which can evade his own meta-classifier well. We will evaluate if this Trojaned model can evade the defender's meta-classifier.

\presubspace
\subsection{Evaluation Results}
\postsubspace

We evaluate \Sys{}-robust over all the Trojaned tasks as in \refsect{perf} and show the results in Table \ref{tab:result-robust}. The first row is the detection performance of \Sys{}-robust on normal Trojaned models (without adaptive attack) and the second row is its detection performance against adaptive attack.

From the first row, we see that the detection performance of our robust \Sys{} does not downgrade much in normal scenario where there is no adaptive attack (\Sys{}-robust versus \Sys{} (Jumbo) in Table~\ref{tab:result-cv}). 
From the second row, we see that the robust version of \Sys{} works much better against adaptive attacks. In some cases the detection performance even increases, which means that the intentional evasion on the attacker-chosen meta-classifier actually makes the Trojaned model easier to detect. By comparison, if we do not add the precautions with randomness, the simple \Sys{} system will be bypassed by all these strong adaptive attacks. 
\presecspace
\section{Discussion \& Limitations}
\postsecspace
\label{sec:diss}
\paragraph{Trojan Attack Detection Levels.}
In this paper, we focus on the model-level Trojan attack detection. Other works may investigate in input-level detection \cite{gao2019strip,chou2018sentinet} or dataset-level detection \cite{chen2018detecting,tran2018spectral}. These are all feasible ways to prevent users from AI Trojans. However, we consider model-level detection the most generally applicable approach. The reason is that dataset-level detection can only detect the Trojans that perform poisoning attack to the dataset. They cannot work against attacks that directly modify model parameters. The input-level detection requires the defender to perform detection each time the input is fed to the model. This will decrease the efficiency when deploying the model. As a comparison, a user only needs to perform model-level Trojan detection one time. As long as no Trojan is detected in the model, the user can deploy it without any cost in the future.

\paragraph{Detection vs. Defense/Mitigation.}
In this paper, we focus on detecting Trojan attacks.
Defense/mitigation and detection on Trojan attacks
are two very related but orthogonal directions.
Existing defense or mitigation approaches perform Trojan removal based on the assumption that the given models are already Trojaned. However, this is problematic in practice as, in most cases, DNN models provided by the model producers are benign. It is unreasonable to perform Trojan removal on benign models which requires extensive computation and time overhead. Moreover, as shown in ~\cite{wang2019neural}, blindly performing mitigation operations can result in substantial degradation in the model's prediction accuracy for benign inputs. Therefore, Trojan detection should be considered as a prerequisite before conducting Trojan mitigation.
Once a model is identified as Trojaned model, the mitigation can be executed more confidently to avoid
a waste of computation and time.

\paragraph{Meta-classifier on other ML models} In the paper, we mainly detect AI Trojans on neural networks. We do not include other ML models in our discussion mainly because there is no current research showing that they suffer from backdoor attacks. We emphasize that our technique can be applied to any differentiable ML models that contains a numerical logit vector in its calculation.

\comment{
\paragraph{Differences with Adversarial Examples.}
Both Trojan attacks and adversarial examples can cause misclassification by the model. However, Trojan attacks provide the adversary with full power over the trigger to generate the misclassifications. The trigger pattern selected by the adversary can work for different inputs.
In contrast, the perturbations made to adversarial examples are specific to the input.
}

\presecspace
\section{Related work}
\label{sec:related}
\postsecspace
\paragraph{Trojan Attacks.}
Several recent research~\cite{liu2017neural, gu2017badnets, trojannn, chen2017targeted, liao2018backdoor, ji2017backdoor} has studied software Trojan attacks on neural networks.
As discussed in \refsect{attacks}, Trojans can be created through poisoning the training dataset or direct manipulation of model parameters. For example, Gu et al.~\cite{gu2017badnets} study backdoor poisoning attacks in an outsourced training scenario where the adversary has full knowledge of the model and training data. Comparably, Chen et al.~\cite{chen2017targeted} also use data poisoning but assume the adversary has no knowledge of the training data and model. On the other hand,~\cite{liu2017neural} directly manipulates the neural network parameters to create a backdoor, while~\cite{trojannn} considers Trojaning a publicly available model using training data generated via reverse engineering. Bagdasaryan et al.~\cite{bagdasaryan2018backdoor} demonstrated that any participant in federated learning can introduce hidden backdoor functionality into the joint global model. 
Besides software Trojans, 
Clements et al.~\cite{clements2018hardware} developed a framework for inserting malicious hardware
Trojans in the implementation of a neural network classifier. 
Li et al.~\cite{li2018hu} proposed a hardware-software collaborative attack framework to inject hidden neural network Trojans.

\paragraph{Trojan Attack Detection.}
Several Trojan attack detection approaches have been proposed~\cite{wang2019neural, gao2019strip, chou2018sentinet, chen2018detecting, ma2019nic}. These approaches can be categorized into input-level detection~\cite{gao2019strip, chou2018sentinet, ma2019nic}, model-level detection~\cite{wang2019neural} and dataset-level detection~\cite{chen2018detecting}.
We discussed the differences of these detection levels in \refsect{diss}. 

\paragraph{Trojan Attack Defense/Mitigation.}
To the best of our knowledge, there are few evaluated defenses against Trojan attacks~\cite{liu2018fine, tran2018spectral}.
Fine-Pruning~\cite{liu2018fine} removes potential Trojans by pruning redundant neurons less useful for normal classification. 
However, the model accuracy degrades substantially after pruning~\cite{wang2019neural}.
The defense in~\cite{tran2018spectral} extracts feature representations of input samples from the later layers of the model and utilizes a robust statistics tool to detect the malicious instances as outliers from each label class.
As discussed in \refsect{diss}, Trojan attack detection and defense are two orthogonal directions. One can first use our approach to detect if a model is Trojaned, then use any of the defenses to remove or mitigate the Trojans.

\paragraph{Poisoning Attacks.}
Poisoning attacks for machine learning models has been well studied in the literature~\cite{biggio2012poisoning, li2016data, munoz2017towards, yang2017generative}.
As discussed in \refsect{attacks}, several Trojan attacks create Trojans through injecting poisoning samples. Those attacks can thus be seen as variants of poisoning attacks.
However, most conventional poisoning attacks seek to degrade a model's classification accuracy on clean inputs~\cite{baracaldo2017mitigating, papernot2016towards}. In contrast, the objective of Trojan attacks is to embed backdoors while not degrading the model's prediction accuracy on clean inputs.

\paragraph{Property Inference.}
Property inference attacks~\cite{ateniese2015hacking, ganju2018property, melis2018inference} aim to infer certain properties about the training dataset or the model of a target model. 
However, as illustrated in \refsect{approach}, detecting Trojaned model using property inference is not a trivial task. We thus propose jumbo learning to construct Trojaned shadow models. Besides, existing work considers white-box access to the target model while we consider black-box access. 
The work of~\cite{melis2018inference} focuses on inference against collaborative learning, which has a different setting as ours. 


\comment{
\paragraph{Watermarking.}
Some work~\cite{adi2018turning} proposed to watermarking deep neural network using backdoors. The argument is that the inserted backdoor can be used to claim the ownership of the model provider since only the provider is supposed to have the knowledge of such backdoor, while the backdoored DNN model has no (or imperceptible) degraded functional performance on normal inputs.
}

\presecspace
\section{Conclusion}
\label{sec:conclusion}

In this paper, we presented \Sys, a novel framework to detect Trojans in neural networks using meta neural analysis techniques.
We propose jumbo learning to generate shadow models without the knowledge of the attacker's approach.
In addition, we provide a comprehensive comparison between existing Trojan detection approaches and ours. We show that \Sys outperforms all the existing detection works in most cases and generalizes well to unforeseen attack strategies. We also design and evaluate a robust version of \Sys{} against strong adaptive attackers. Our work sheds new light on the detection of Trojans in neural networks.

\section*{Acknowledgements}
This material is based upon work supported by the Department of Energy under Award Number DE-OE0000780. The views and opinions of authors expressed herein do not necessarily state or reflect those of the United States Government or any agency thereof.

\bibliographystyle{plain}
\bibliography{ref/ref,ref/thesisrefs.bib}

\appendix
\subsection{Detailed Discussion on Existing Detection Approaches}
\label{sec:app-detect-compare}
We find that existing approaches have different assumptions on the defender and detection capabilities. For example, dataset-level works require access to the training set and cannot detect model manipulation attacks which do not poison the dataset; works inspired by anomaly detection cannot detect all-to-all attacks where all the labels suffer from the same level of attack so that there is no anomaly. In the following we introduce the assumptions and capabilities of existing detection approaches.

Neural Cleanse(NC)~\cite{wang2019neural} and DeepInspect(DI)~\cite{chen2019deepinspect} work on the same level with us. NC observes that in a Trojaned model, there exists a short-cut modification (i.e. the trigger pattern) to change any input to be predicted as the Trojan label. Therefore, it calculates such modification for each label and checks if there exists a short-cut which is much smaller in size than the modifications of other labels.
DI improves upon the approach by using model inversion to get some training data. Then they use GAN to generate the modifications and apply the same algorithm to check short-cut as in NC. Both approaches cannot be applied to detect all-to-all attack where the pattern itself cannot lead to certain Trojan label, so the short-cut no longer exists. They cannot be applied to detect Trojans in binary classification tasks because their short-cut check algorithm requires at least three classes. In addition, NC performs not well in detecting large-size triggers~\cite{chen2019deepinspect}.

Activation Clustering (AC)~\cite{chen2018detecting} and Spectral Signature~\cite{tran2018spectral} work on the dataset-level detection. AC performs a two-class clustering over the feature vector of the training data to separate benign data and Trojaned data (if exists). Spectral calculates a signature score for each data in the training set to remove the ones which possibly contain a Trojan trigger. These approaches perform detection on the dataset level, so they need access to the training data and cannot be applied to detect model manipulation attacks which do not poison the dataset. AC also requires white-box access to calculate the feature vector.

STRIP~\cite{gao2019strip} and SentiNet~\cite{chou2018sentinet} detects Trojans on the input level. STRIP adds up the input with other clean data. The network will give a confident answer on the mixed input if it contains the Trojan pattern; otherwise the network will be confused. SentiNet uses computer vision techniques to find salient parts in the image, which are possibly the Trojan trigger pattern. It then copies the parts to other images to check if it can change the output of other images. Both approaches need a set of clean data to detect Trojans. STRIP cannot detect all-to-all attacks where the model cannot give a confident answer even if it sees the trigger pattern on the blended input. SentiNet requires white-box access to detect salient part and cannot detect large-size trigger via saliency check.

\subsection{One-Class SVM}
\begin{figure}[tbp]
    \centering
    \includegraphics[width=0.35\textwidth]{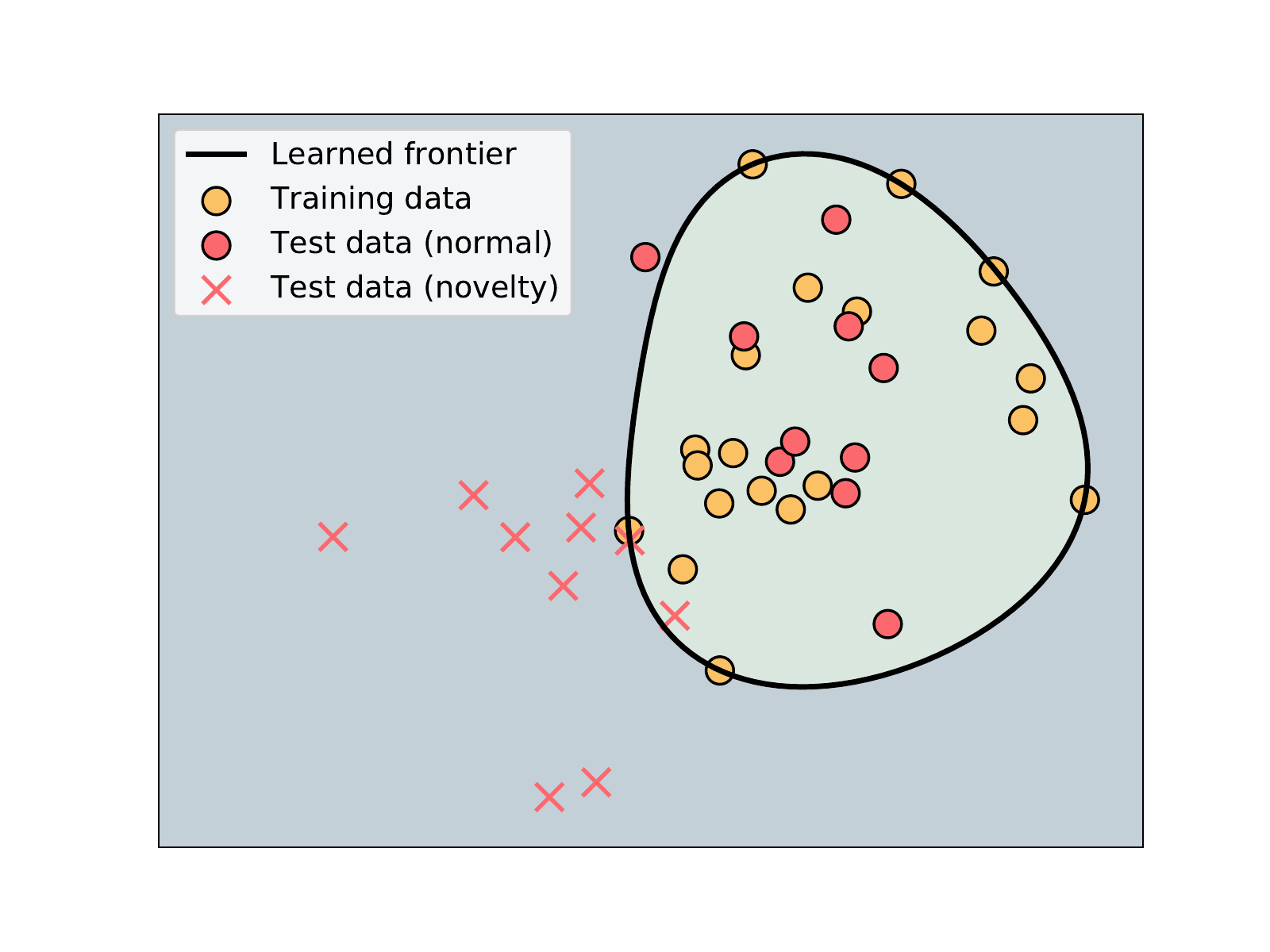}
    \caption{An illustration of the idea of one-class SVM.}
    \label{fig:ocsvm}
\end{figure}

In \reffig{ocsvm}, we illustrate an example of the one-class SVM model. The model is only provided with a set of training data in one class, and tries to find the decision boundary that captures all the training data tightly. The test set consists of data in the class and data outside the class, and the goal is to distinguish between the two classes.

\label{app:svm}



\subsection{Dataset Details and Network Structures}
\label{app:structure}

\begin{table}[tbp]
    \centering
    \footnotesize
    \caption{The model structure for each dataset in our evaluation. Each convolutional layer and linear layer are followed by a ReLU activation function except the last linear layer.}
    \begin{tabular}{c|c}
        \hline
        \textbf{\mnist} & \textbf{\cifar} \\
        \hline
        Conv ($16\times5\times5$, pad=0) & Conv ($32\times3\times3$, pad=1)\\
        MaxPool ($2\times2$) & Conv ($32\times3\times3$, pad=1)\\
        Conv ($32\times5\times5$, pad=0) & MaxPool ($2\times2$)\\
        MaxPool ($2\times2$) & Conv ($64\times3\times3$, pad=1)\\
        Linear ($512$) & Conv ($64\times3\times3$, pad=1)\\
        Linear($10$) & MaxPool ($2\times2$)\\
        & Linear ($256$)\\ 
        & Linear ($256$)\\ 
        & Dropout ($0.5$)\\ 
        & Linear ($10$)\\ 
        \hline
        \hline
        \textbf{\irish} & \textbf{\nlp} \\
        \hline
        LSTM (100, layer=2) & Word Embedding (300)\\
        Attention & Conv ($100\times\{3,4,5\}\times300$)\\
        Linear (1) & Concatenation\\
         & Dropout (0.5)\\
         & Linear(1)\\
        \hline
        \hline
        \textbf{\speech} \\
        \hline
        MelSpectrogram Extraction\\
        LSTM (100, layer=2)\\
        Attention\\
        Linear ($10$)\\
        \hline
    \end{tabular}
    \label{table:model-structure}
\end{table}

\paragraph{Computer Vision.} 
We use the standard MNIST~\cite{lecun_cortes_burges} and CIFAR10~\cite{cifar} datasets for computer vision tasks. 
\mnist contains 70,000 handwritten digits with 60,000 samples used for training and 10,000 samples for testing. Each data
sample is a 28x28 grayscale image.
\cifar consists of 60,000 32x32 RGB images in 10 classes, with 50,000 images for training and 10,000 images for testing.
For \mnist, we adopt the same CNN structure as in~\cite{gu2017badnets}. For \cifar, we use the same CNN structure as in \cite{carlini2017towards}. 

\paragraph{Speech.} We use the SpeechCommand dataset (\speech) version~v0.02 \cite{warden2018speech} for the speech task. 
The \speech dataset consists of 65,000 audio files, each of which is a one-second audio file belonging to one of 35 commands. We use the files of ten classes (``yes'', ``no'', ``up'', ``down'', ``left'', ``right'', ``on'', ``off'', ``stop'', ``go'') as \cite{googlespeechcommand} does and it gives 30,769 training samples and 4,074 testing samples. Given the audio signal files, we first extract the mel-spectrogram of each file with 40 mel-bands. Then we train an Long-Short-Term-Memory (LSTM) network over all the mel-spectrograms.

\paragraph{Tabular Records.} 
We use the Smart Meter Electricity Trial data in Ireland dataset (\irish)~\cite{irish} for tabular data tasks.
The \irish dataset consists of the electricity consumption of 4,710 users in 76 weeks.
Each record has 25,536 columns with each column being the electricity consumption (in kWh) of users during 30 minute intervals.
Each user is labeled as residential or SME (Small to Medium Enterprise).
We split the dataset to have 3,768 users (80\% of all users) in the training set and 942 (20\%) in the test set.
For the training set we use the data in the first 46 weeks (60\% of the total time length) while for the test set we use the data in the last 30 weeks (40\%). 
We use the electricity consumption in each week as the feature vector and view the vectors of all the weeks as a time series. Then we train an LSTM model to predict whether a given electricity consumption record belongs to a residential user or an SME.

\paragraph{Natural Language.} 
We use the same Rotten Tomatoes movie review dataset (\nlp)\ as Kim~\cite{kim2014convolutional} for natural language processing tasks. 
The \nlp dataset consists of 10,662 movie review sentences. The task is to determine whether a movie review is positive or negative. 
Following the convention of the previous work~\cite{kim2014convolutional}, we use 90\% of the data for training and the rest for testing. We use the same model structure as Kim~\cite{kim2014convolutional} except that we use a pretrained and fixed Gensim model as the word embedding layer. A pretrained embedding layer provides a better performance given the limited training data we use.

For reproduction, the model structures for the evaluation on each dataset are presented in Table~\ref{table:model-structure}. The hyperparameters of the layers are shown in the parenthesis following the layer name. For convolutional layers, the number of filters, filter width and filter height, as well as the padding are listed. 
For linear layers, we omit the input size and only show the output size.





\subsection{Defense on Discrete Data}
\label{sec:app-discrete}
On the \nlp task, the input words are in discrete token space. Therefore, we cannot use gradient-based approach to do query-tuning. 
However, for most neural networks with discrete input space, the input will first be mapped to some continuous embedding space (e.g., word2vec in NLP). Thus, we will optimize the ``query set'' over the embedding space to in the same way as before. During inference, we directly feed the tuned embedding vectors to the target model to get predictions. The trade-off is that under this setting, we need white-box access to the embedding layer of the target model. We adopt this setting in the NLP tasks.

\subsection{Detection Baselines Implementation Details}
\label{sec:app-detect-baseline}
At the time of writing, only the source code of NC is released.
Moreover, all the baselines only evaluate with CNN models on computer vision datasets in their work, except for AC where CNN models on NLP dataset are also evaluated.
To compare our approaches with these baselines, we re-implement them with Pytorch.

Since AC and Spectral are dataset-level detection and STRIP is input-level detection, we will tailor them to detect model-level Trojans to compare them with our pipeline. AC works on the dataset level and uses an \emph{ExRe} score to indicate whether the dataset is Trojaned. We use this score to indicate the Trojan score for the model. Spectral assigns a score to each training sample. We use the average score of all the training data to indicate the score of the model being Trojaned. STRIP predicts whether an input data is Trojaned, we use their approach to calculate a score for each training sample and take the average to indicate the likelihood of a Trojaned model.

\begin{figure}[t]
\vspace{-1em}
    \centering
    \includegraphics[width=0.23\textwidth]{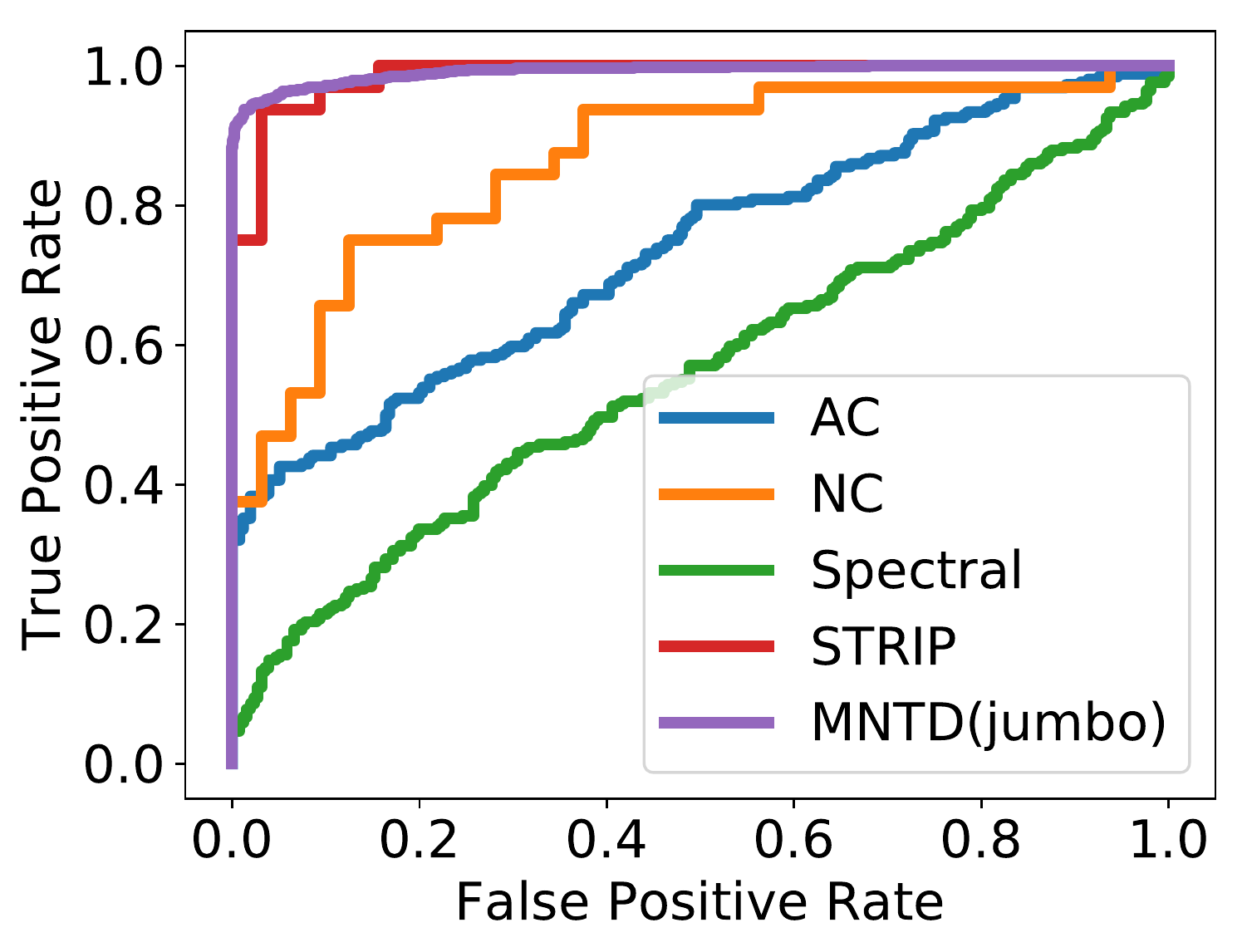}
    \includegraphics[width=0.23\textwidth]{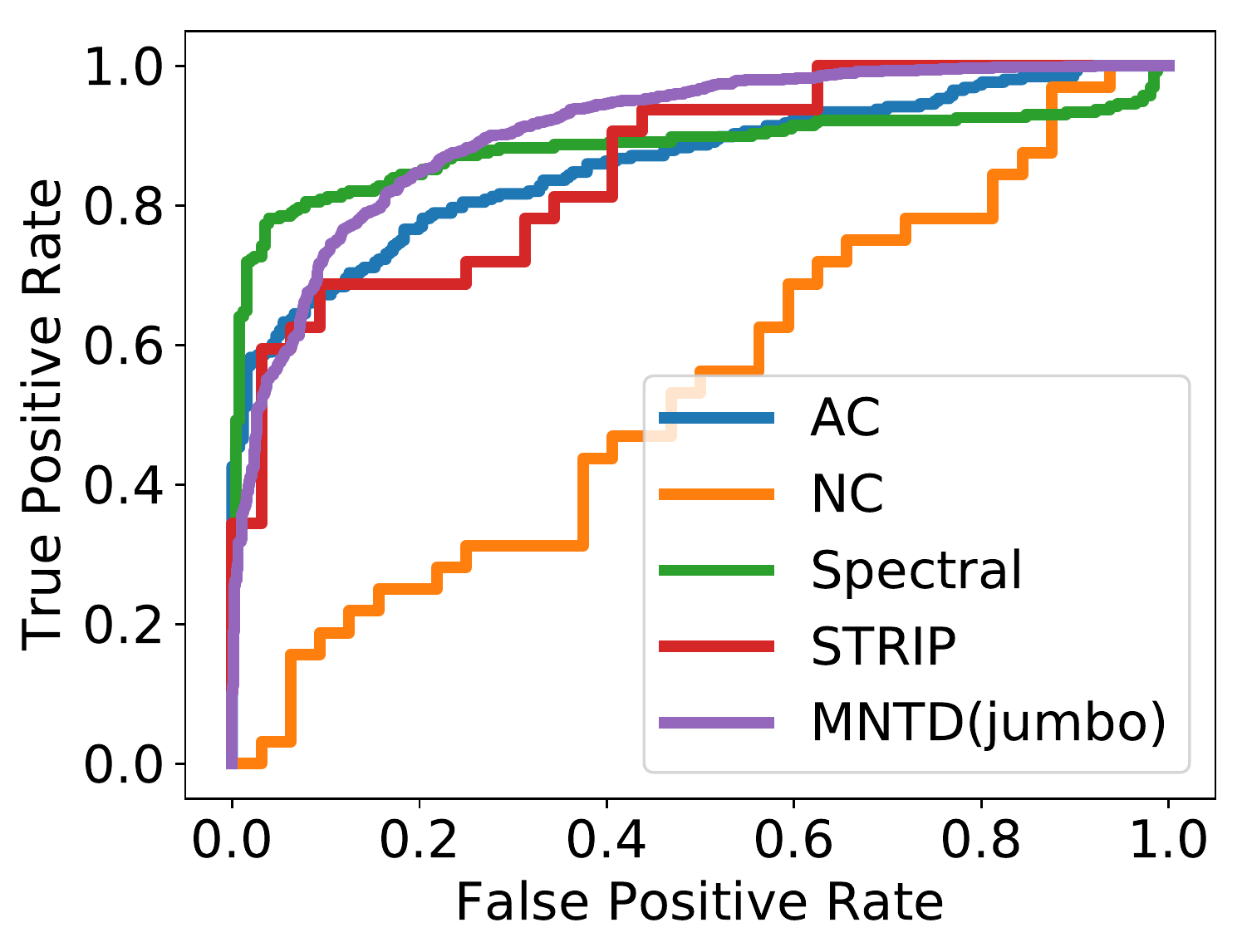}
    \caption{The detection ROC curve of different approaches on \mnistm (left) and \cifarm(right).}
    \label{fig:roc}
\vspace{-1em}
\end{figure}


\begin{figure*}[htbp]
    \centering
    \includegraphics[width=0.9\textwidth]{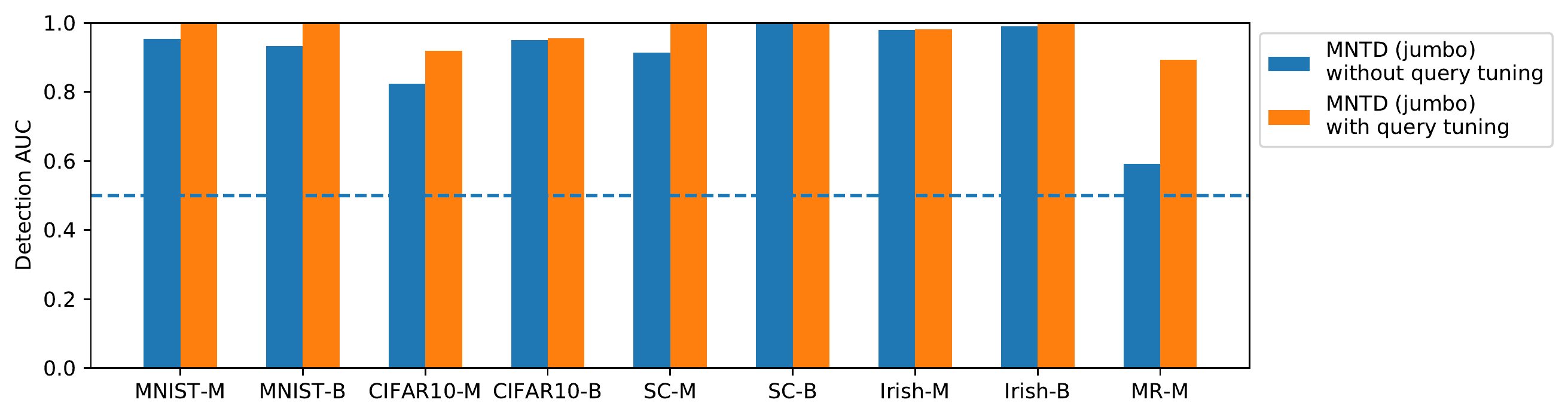}
    \caption{The comparison of detection AUC with and without query tuning. -M stands for modification attack and -B stands for blending attack.}
    \label{fig:result-querytune}
    \vspace{-1.5em}
\end{figure*}

\subsection{Other Experiment Results}
\label{sec:app-other-exp}
\paragraph{ROC Curve of detection} We show the ROC curve of the detection performance for different approaches on \mnistm and \cifarm as in Figure~\ref{fig:roc}.

\paragraph{Effectiveness of Query Tuning}

We compare the results of Jumbo \Sys{} with and without query tuning in Figure~\ref{fig:result-querytune}.
The results show that query tuning is highly effective; 
the AUC scores drop as much as 30\% in the worst case if we use untuned queries instead.
We can interpret the improvement by an analogy to feature engineering: we would like to obtain shadow model features with the most distinguishability for the meta-model to do classification.
In query-tuning the feature engineering is done by tuning the queries.

\begin{figure}
    \centering
    \begin{subfigure}[t]{0.23\textwidth}
        \centering
        \includegraphics[width=\linewidth]{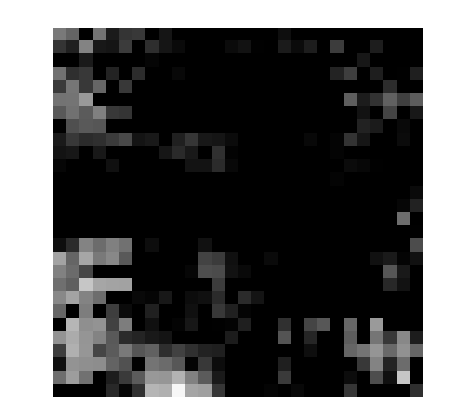}
        \caption{Jumbo}
        \label{fig:queries-c}
    \end{subfigure}
    \begin{subfigure}[t]{0.23\textwidth}
        \centering
        \includegraphics[width=\linewidth]{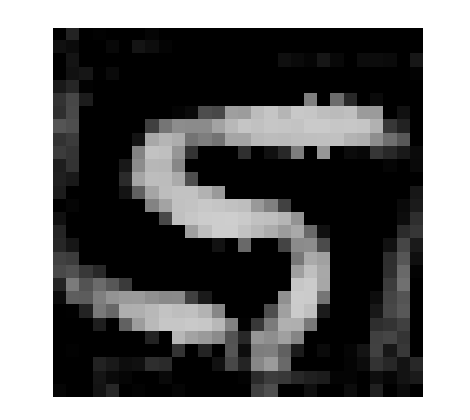}
        \caption{One-class}
        \label{fig:queries-d}
    \end{subfigure}
    \caption{Example of tuned-queries in MNIST. To make the pattern more clear, we magnify the contrast of the jumbo query by 5 times.}
    \label{fig:queries}
\vspace{-1.5em}
\end{figure}

We show some of the tuned queries on the \mnistm task in Figure~\ref{fig:queries}.
We observe that the tuned query in jumbo learning focuses more on local patterns, while the tuned query in one-class learning contains more global and digit-like pattern.
We speculate that it is because most Trojaned models in jumbo learning use small local pattern, so this query can help distinguish between benign model and jumbo Trojaned models. On the other hand, the one-class learning needs to fit the benign models best, so the query looks like normal benign input.

\begin{figure}[t]
    \vspace{-1em}
    \centering
    \begin{subfigure}[b]{0.24\textwidth}
    \centering
        \includegraphics[height=1.8cm]{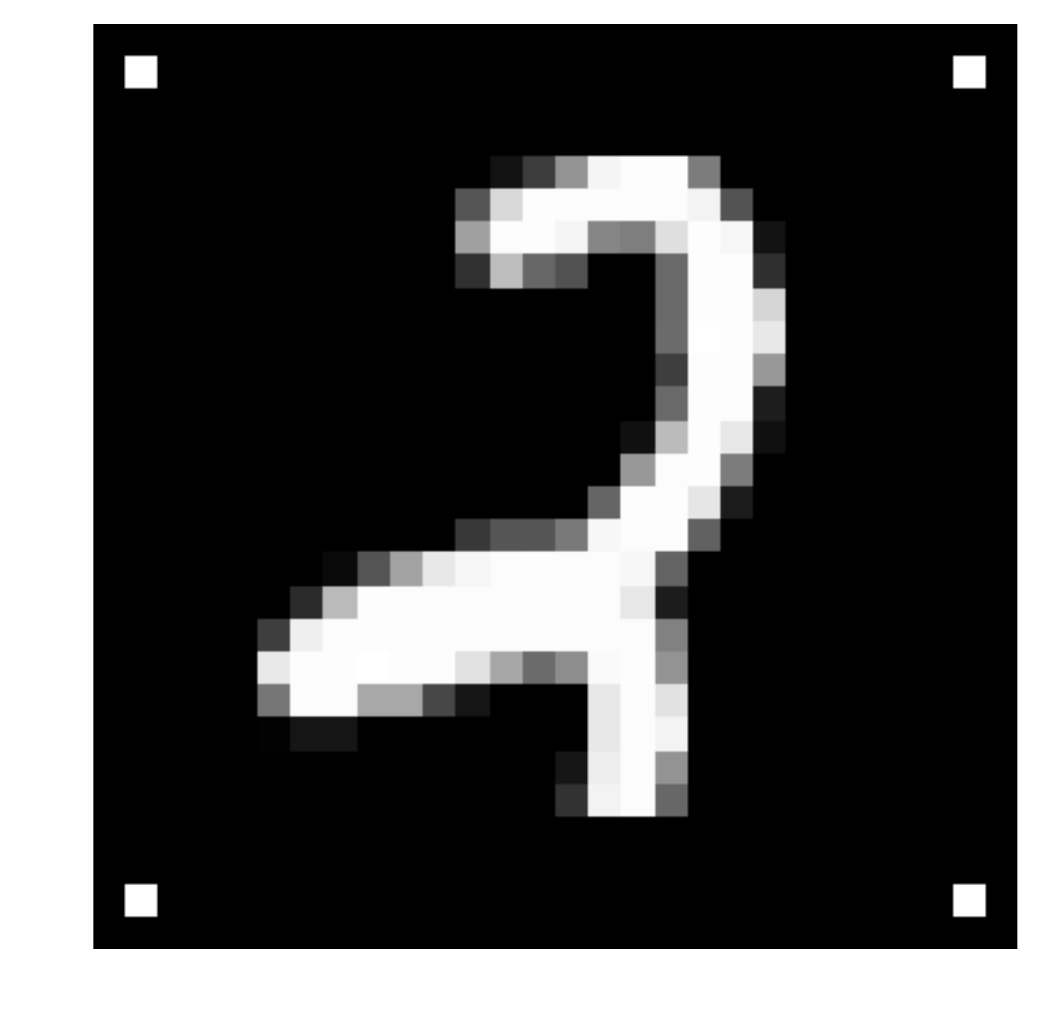}
        \includegraphics[height=1.8cm]{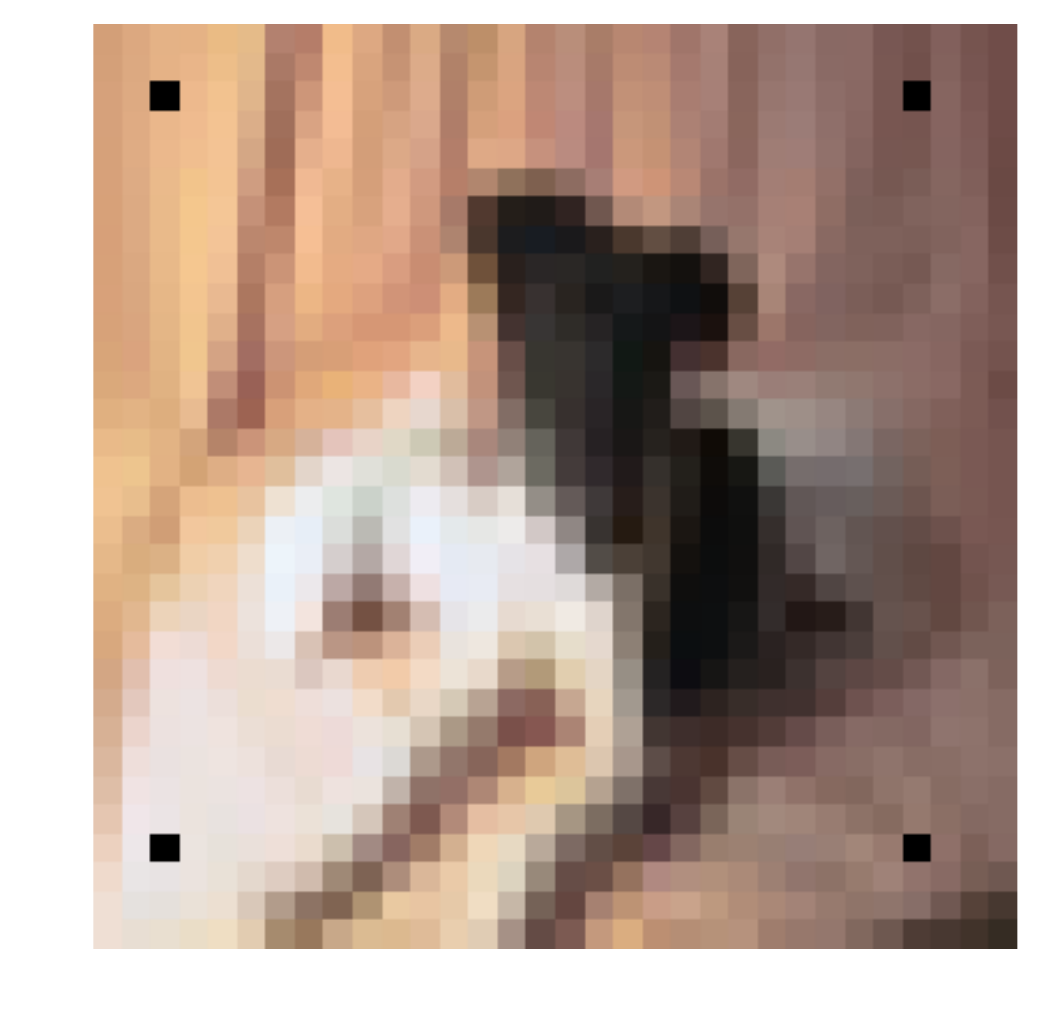}
        \caption{Test patterns.}
        \label{fig:mnist-nv-eg}
    \end{subfigure}
    \begin{subfigure}[b]{0.24\textwidth}
    \centering
        \includegraphics[height=1.8cm]{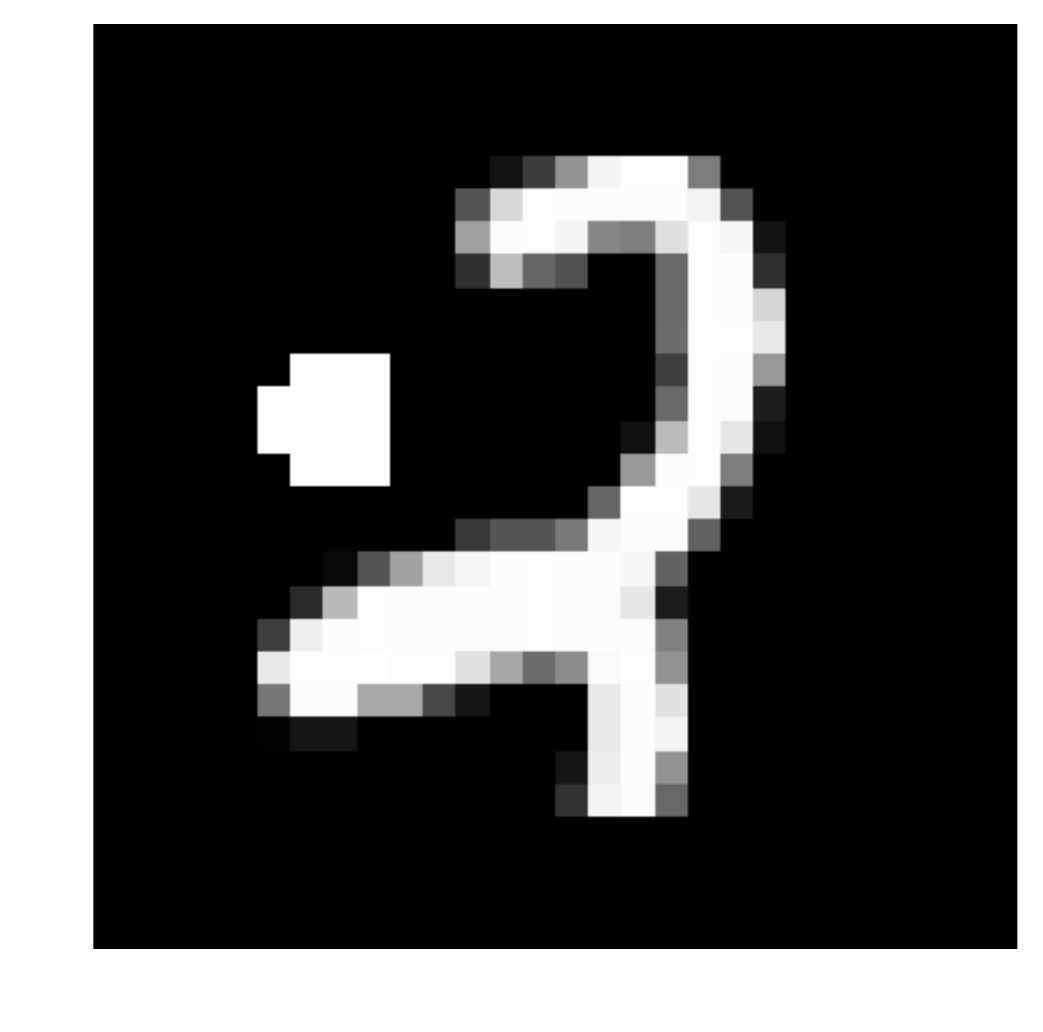}
        \includegraphics[height=1.8cm]{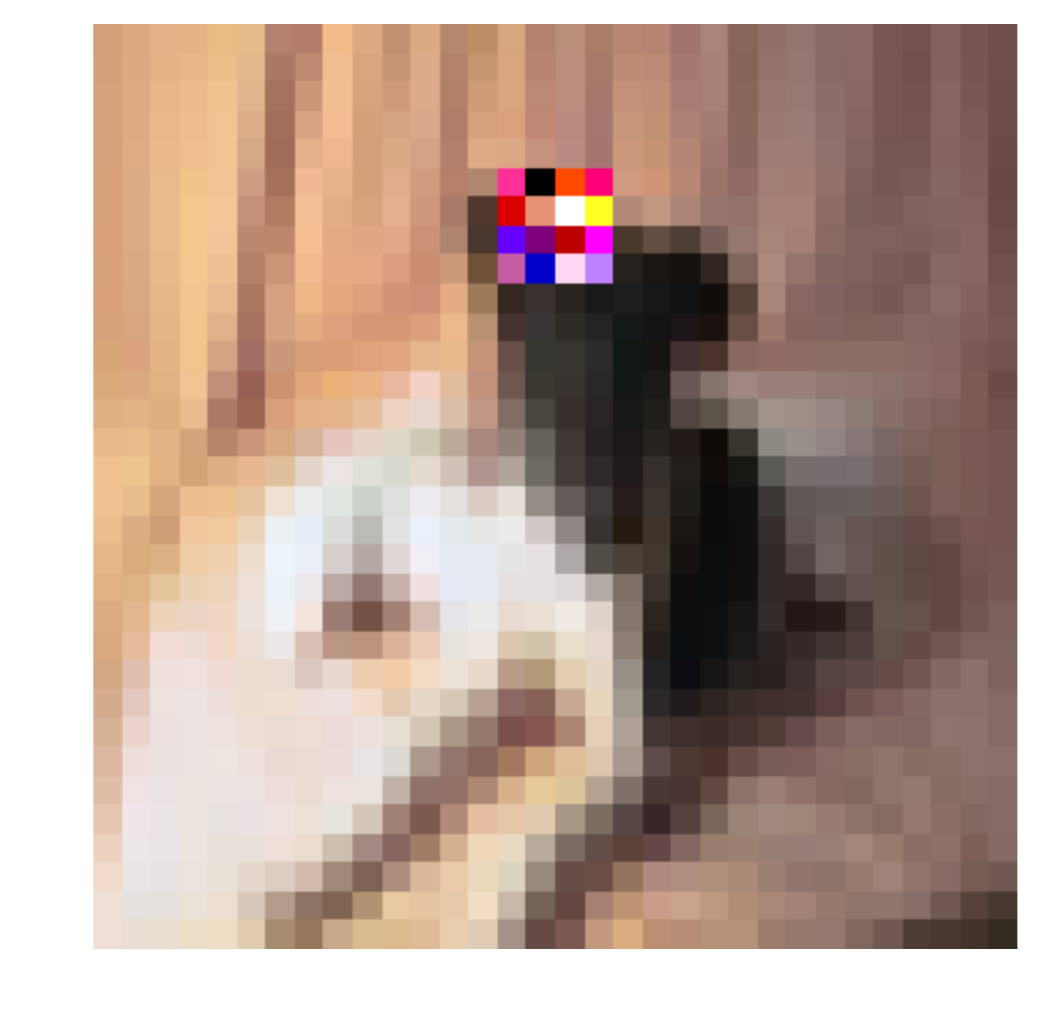}
        \caption{Examples of training patterns.}
        \label{fig:mnist-nonnv-eg}
    \end{subfigure}
    \caption{The unforeseen trigger pattern used in evaluation (left) and examples of the trigger patterns used in jumbo training (right) on MNIST and CIFAR.}
    \label{fig:generalize-pattern}
    \vspace{-0.5em}
\end{figure}

\paragraph{Generalization on Trigger Patterns} In the jumbo distribution we assume that the trigger patterns are all consecutive patterns (e.g., one square pattern in vision task). Now we will evaluate the meta-classifier using Trojans with non-consecutive patterns. These patterns will never appear in the jumbo distribution. In particular, we modify one pixel at each of the four corners and use it as the trigger pattern for vision tasks (see Figure \ref{fig:mnist-nv-eg}). For \speech, we modify the signal value in the first 0.25 second, middle 0.5 second and last 0.25 second to be 0.1. For \irish, we modify the usage from 9:00 am to 10:00 am on every weekday to be 0. For \nlp, we add a ``yes'' at the beginning and an ``oh'' at the end of the sentence.

The results are shown in Table \ref{tab:result-unseen-pattern}. We see that the meta-classifier achieves the similar performance as when detecting the triggers that we have seen. This shows that the trained meta-classifier generalizes well even if the trigger patterns are not considered in the training process.

\begin{table}[t]
    \centering
    \caption{The detection AUC of jumbo \Sys{} on Trojaned models with unforeseen trigger pattern.}
     \footnotesize
    \label{tab:result-unseen-pattern}
    \begin{tabular}{c|c|c|c|c}
        \hline
         \mnist & \cifar & \speech & \irish & \nlp \\
        \hline
         100.00\% & 96.97\% & 93.21\% & 100.00\% & 94.32\% \\
        \hline
    \end{tabular}
\vspace{-1em}
\end{table}

\subsection{Label mapping of TinyImageNet}
\label{sec:app-label-map}
The classes we choose in TinyImageNet which correspond to the labels in CIFAR-10 are shown in Table~\ref{tab:tinyimagenet-class}. Note that, the `airplane' class corresponds to the images of rockets and `horse' corresponds to the images of camels, since the TinyImageNet does not contain images of airplanes and horses.

\begin{table}[t]
    \centering
    \caption{The classes in CIFAR-10 and corresponding class picked in TinyImageNet.}
     \footnotesize
    \label{tab:tinyimagenet-class}
    \begin{tabular}{c|c}
        \hline
        class in CIFAR-10 & class in TinyImageNet \\
        \hline
        airplane & n04008634\\
        automobile & n02814533\\
        bird & n02002724\\
        cat & n02123045\\
        deer & n02423022\\
        dog & n02085620\\
        frog & n01641577\\
        horse & n02437312\\
        ship & n03662601\\
        truck & n03796401\\
        \hline
    \end{tabular}
\vspace{-1em}
\end{table}

\end{document}